\documentclass[review]{elsarticle}

\usepackage{lineno,hyperref}

\usepackage{amsmath}
\usepackage{amsfonts}
\usepackage{graphicx}
\usepackage{multirow}
\usepackage[lofdepth,lotdepth]{subfig}
\usepackage{float}

\usepackage{enumerate}
\usepackage{booktabs}

\usepackage[tableposition=top]{caption}

\biboptions{authoryear}

\begin{document}

\begin{frontmatter}

\title{Multi-View Fuzzy Clustering with Minimax Optimization for Effective Clustering of Data from Multiple Sources}


\author[mymainaddress]{Yangtao Wang\corref{mycorrespondingauthor}}
\cortext[mycorrespondingauthor]{Corresponding author}
\ead{wangyt@i2r.a-star.edu.sg}

\author[mysecondaryaddress]{Lihui Chen}
\ead{ELHCHEN@ntu.edu.sg}
\address[mymainaddress]{Connexis(South Tower), Data Analytics Department, Institute for Infocomm Research, A*STAR, 1 Fusionopolis Way, Singapore 138632, Republic of Singapore}
\address[mysecondaryaddress]{ INFINITUS Centre for Infocomm Technology, School of Electrical and Electronic Engineering, Nanyang Technological University, 50 Nanyang Ave, Singapore 639798, Republic of Singapore}

\begin{abstract}
Multi-view data clustering refers to categorizing a data set by making good use of related information from multiple representations of the data. It becomes important nowadays because more and more data can be collected in a variety of ways, in different settings and from different sources, so each data set can be represented by different sets of features to form different views of it.  Many approaches have been proposed to improve clustering performance by exploring and integrating heterogeneous information underlying different views.  In this paper, we propose a new multi-view fuzzy clustering approach called MinimaxFCM by using minimax optimization based on well-known Fuzzy c means. In MinimaxFCM the consensus clustering results are generated based on minimax optimization in which the maximum disagreements of  different weighted views are minimized. Moreover, the weight of each view can be learned automatically in the clustering process. In addition, there is only one parameter to be set besides the fuzzifier. The detailed problem formulation, updating rules derivation, and the in-depth analysis of the proposed MinimaxFCM are provided here. Experimental studies on nine multi-view data sets including real world image and document data sets have been conducted.
We observed that MinimaxFCM outperforms related multi-view clustering approaches in terms of clustering accuracy, demonstrating the great potential of MinimaxFCM for multi-view data analysis.
\end{abstract}


\end{frontmatter}


\section{Introduction}

%
%

Multi-view data becomes common nowadays because data can be collected from different sources or represented by different features. For example, the same news can be reported in different articles from different news sources, one document can be translated into different kinds of languages and one image can be represented with different kinds of features. Learning and analyzing multi-view data has become a hot research topic in recent years and attracted many researchers in, to name a few, the areas of data mining, machine learning, information retrieval and cybersecurity. Many multi-view learning approaches based on different strategies including co-training \citep{blum1998combining}, multiple kernel learning \citep{lanckriet2004learning} and subspace learning \citep{jia2010factorized} have been proposed in the literature \citep{xu2013survey}. In \citep{xu2015multi} a multi-view learning approach based on subspace learning was proposed to discover a latent intact representation of the data. In \citep{wang2015deep} deep neural networks were used to learn representations (features) for multi-view data. Multi-view learning approaches can be divided into supervised learning and unsupervised learning approaches. In this paper, we focus on one of the unsupervised learning techniques which is clustering for multi-view data analysis. As a promising data analysis tool, clustering is able to find the pattern structure and information underlying the unlabelled data. Clustering algorithms based on different theories have been proposed in various applications in the literature \citep{jain2010data,filippone2008survey,xu2005survey}.
Multi-view clustering approaches are able to mine valuable information underlying different views of data and integrate them to improve clustering performance which have wild applications. For example, in news articles categorization, each article may be  written in different languages or collected from different news sources. In an e-learning education system, students' behaviour and performance in study may be analysed based on some features collected from various sources. Students may be clustered into different groups based on several sets of features, for example, how they approach the exercises, and how they interact with the tutorial videos, those form two different sets of features.

Many multi-view clustering approaches have been proposed  in the literature. For clustering multi-view data, roughly three strategies are applied among the existing approaches. The first strategy is to integrate multi-view data into a single objective function which is optimized directly during the clustering process. The consensus clustering result is generated directly without one more step to combine the clustering result of each view.  For example, in \citep{2011NIPScosc}, two co-regularized multi-view spectral clustering algorithms were proposed. The pairwise disagreement term and centroid based disagreement term for different views are added into the objective function of spectral clustering. The clustering results which are consistent across the views are achieved after the optimization process. In \citep{2012ICDMmvKernelKmeans}, a kernel-based weighted multi-view clustering approach was presented. In particular, each view is expressed by a kernel matrix. The weight of each view and consensus clustering result are learned by minimizing the disagreements of different views. In \citep{2013IJCAIRmultiviewKmeans}, a multi-view clustering approach based on K-means was proposed. The consensus cluster indication is integrated in the objective function directly. The second strategy includes two steps as follows. First, a unified representation (view) is generated based on multiple views. Then the existing clustering algorithm such as K-means \citep{macqueen1967some} or spectral clustering \citep{ng2002spectral} is applied to achieve the final clustering result. For example, in \citep{2012cvpraffinity}, Huang et al. propose an affinity aggregation spectral clustering in which an aggregated affinity matrix is found first by seeking the optimal combination of different affinity matrices. Then spectral clustering is applied on the new affinity matrix to get the final clustering result. In \citep{2013AAAIconvexSubspaceMultiview}, a common subspace representation of the data shared across multiple views is first learned. Then K-means is applied on the learned subspace representation matrix to generate the clustering result. In the third strategy, each view of the data is processed independently and an additional step is needed to generate the consensus clustering result based on the result of each view. For example, in \citep{bruno2009multiview} and \citep{greene2009matrix}, the consensus clustering result was achieved by integrating the previously generated clusters of individual views based on the latent modeling of cluster-cluster relationships and matrix factorization respectively.

The above multi-view clustering approaches are all based on hard clustering in which each object can only belong to one cluster. Since the real world data sets may not be well separated, different approaches have been proposed based on soft or fuzzy clustering algorithms \citep{aparajeeta2016modified,kannan2015robust,anderson2013comparing}
 in which each object can belong to all the clusters with various degrees of memberships. The memberships used in soft clustering help to describe the data better and have many potential applications in the real world. For example, soft clustering approaches can better capture the topics of each document which  belongs to several topics with different degrees. In \citep{liu2013multi}, Liu et al. propose a joint Nonnegative Matrix Factorization (NMF) \citep{lee1999learning} approach for multi-view clustering in which a disagreement term is introduced in the objective function. Besides NMF based multi-view clustering approaches, several multi-view fuzzy clustering algorithms based on the well known Fuzzy c means (FCM) algorithm \citep{Bezdek:1981:PRF:539444} have been developed. For example, in \citep{2009COFKM}, CoFKM is proposed to handle multi-view data by minimizing the objective function of FCM of each view and penalizing the disagreement between any pairs of views. In \citep{2014WVCOFCM}, a multi-view fuzzy clustering with weighted views called WV-Co-FCM was proposed. In WV-Co-FCM, the clustering process is based on optimizing the objective function which highlights the fuzzy partition and the weight of each view is achieved by introducing the entropy regularization term.

Both hard and soft approaches discussed above all formulate the multi-view clustering to an optimization problem in which  the disagreement of the views is minimized. In \citep{wang2014multi}, a minimax optimization based multi-view spectral clustering approach was proposed to handle multi-view relational data.
However, as pointed out in \citep{2013IJCAIRmultiviewKmeans}, the spectral clustering based multi-view clustering approaches have two drawbacks. One is that the clustering performance is sensitive to the choice of the kernel to build the graph. The other is that they are not suitable for large scale data clustering because of the high time computational cost on kernel construction as well as eigen decomposition. Fuzzy c means (FCM) is widely applied in many applications
because of its effectiveness and low time complexity. To combine the advantages of minimax optimization and FCM, in this paper we propose MinimaxFCM  for multi-view data clustering. In MinimaxFCM, the goal is to achieve the consensus clustering result of multi-view data by minimizing the maximum disagreement of the weighted views. Except for the fuzzifier which is one parameter in all FCM based approaches, there is only one extra parameter in MinimaxFCM to control the distribution of each view. Moreover, the time complexity of MinimaxFCM is similar to FCM. 
 The experiments with MinimaxFCM on nine real world data sets including image and document data sets show that MinimaxFCM achieves better clustering performance than the related clustering approaches.

The rest of the paper is organized as follows: in the next section, the highlights of the related multi-view clustering approaches reported in the literature are given. In Section III, the details of the proposed multi-view fuzzy clustering approach MinimaxFCM are described. Experiments on the real world data sets are conducted and the results are analyzed in Section IV. Finally, conclusions are drawn in Section V.

\section{Related work}

In this section, five related multi-view clustering approaches including two hard clustering approaches and three soft clustering approaches are reviewed. Two hard clustering approaches are a K-means based multi-view clustering and the minimax optimization based multi-view spectral clustering. Three soft approaches include one Nonnegative Matrix Factorization based approach and two fuzzy clustering based approaches are reviewed.
\subsection{Notations}
Throughout this paper, the following notations are used unless otherwise stated: we denote the data set which has N objects and K classes as $X=\{x_{1},...x_{N}\}$. The data set is represented by $P$ different views such that the $i_{th}$ object in $p_{th}$ view is denoted as $x_{i}^{p}$. We use $u_{ci}^{p}$ to denote the fuzzy membership which represents the degree of object $i$ belongs to cluster $c$ in $p_{th}$ view and $u_{ci}^{\ast}$ to denote the consensus membership of object $i$ to cluster $c$ shared across different views. The centroid of cluster $c$ of the $p_{th}$ view is denoted as $v_{c}^{(p)}$.  $d_{ic}^{p}=\|x_{i}^{(p)}-v_{c}^{(p)}\|$ is used to denote the distance between centroid $v_{c}^{(p)}$ and object $i$ in $p_{th}$ view and $m$ is used to denote the fuzzifier.
\subsection{RMKMC}
RMKMC \citep{2013IJCAIRmultiviewKmeans} is a multi-view clustering approach based on K-means. The first strategy as discussed in section I is used by RMKMC in which a single objective function is formulated and the consensus clustering result is generated directly after the algorithm converges. In RMKMC, the objective function of K-means is reformulated based on the fact that G-orthogonal non-negative matrix factorization (NMF) is equivalent to relaxed K-means clustering \citep{ding2005nonnegative}. To make the approach more robust to outliers, the $l_{2,1}$ norm is applied in the objective function as follows.
\begin{equation}
 \min\limits_{G^{\ast}, \{F^{(p)}\}_{p=1}^{P},\{\alpha^{(p)}\}_{p=1}^{P}}\;\sum\limits_{p=1}^{P}(\alpha^{(p)})^{\gamma}\|(X^{(p)})^{T}-G^{\ast}(F^{(p)})^{T}\|_{2,1}
\label{equation:JRMKMC}
\end{equation}
Here $G$ is the coefficient matrix which is considered as the cluster indicator matrix. $F^{(p)}$ is the basis matrix of the $p_{th}$ view which can be considered as the cluster centroid matrix. As shown in the objective function, the summation of the weighted difference of each view is minimized and the consensus clustering result $G^{\ast}$ is achieved directly after the algorithm converges. Moreover, the weight of each view $\alpha^{(p)}$ is updated automatically. The higher the value of the weight, the more important the view is.

\subsection{MinimaxMVSC}
MinimaxMVSC \citep{wang2014multi} is a multi-view spectral clustering approach based on minimax optimization. In MinimaxMVSC, the first strategy is used to formulate the objective function as follows:
\begin{equation}
 \min\limits_{I^{\ast}, \{M^{(p)}\}_{p=1}^{P}}\quad\max\limits_{ \{\alpha_{ij}\}_{j\geq i}^{P}}\;\sum\limits_{i=1}^{P}\sum\limits_{j=1}^{P}(\alpha_{ij})^{\gamma}Q_{ij}
\label{equation:MinimaxMVSC}
\end{equation}
Where $I^{\ast}$ is the final consensus cluster indicator matrix and $M^{(p)}$ is the Laplacian embedding of $p_{th}$ view. $Q_{ii}$ is the standard objective function of spectral clustering of $i_{th}$ view and $Q_{ij}$ measures the disagreement of $i_{th}$ view and $j_{th}$ view. The aim of MinimaxMVSC is to minimize the maximum summation of $Q_{ij}$ weighted by $\alpha_{ij}$ to achieve the consensus cluster indicators matrix $I^{\ast}$. Then K-means is applied on $I^{\ast}$ to get the final clustering results.
\subsection{MultiNMF}
In \citep{liu2013multi}, the multiview clustering based joint Nonnegative Matrix Factorization (MultiNMF) is proposed. In MultiNMF, using the first strategy the objective function of joint nonnegative matrix factorization is formulated as follows to find the consensus clustering result.
\begin{equation}
 \sum\limits_{p=1}^{P}\|(X^{(p)})^{T}-G^{(p)}(F^{(p)})^{T}\|_{F}^{2}+\sum\limits_{p=1}^{P}(\lambda^{(p)})\|G^{(p)}-G^{\ast}\|_{F}^{2}
\label{equation:JMNMF}
\end{equation}
Where $\|\cdot\|_{F}$ is the Frobenius norm and $G^{(p)}, F^{(p)}, G^{\ast} \geq 0$. The first term measures the standard NMF reconstruction error for individual views. The second term measures the disagreement between each cluster indicator matrix $G^{(p)}$ and the consensus cluster indicator matrix $G^{\ast}$. The parameter $\lambda^{(p)}$ is set by the user to control the relative weight among different views and between the two terms. To keep the disagreement across different views $\|G^{(p)}-G^{\ast}\|_{F}^{2}$ meaningful and comparable, a novel normalization strategy was proposed by exploring the relation between NMF and Probabilistic Latent Semantic Analysis (PLSA) \citep{hofmann1999probabilistic}. Specifically, $l_{1}$ normalization is conducted with respect to the basis vectors in $F^{(p)}$ during the optimization.
\subsection{CoFKM}
CoFKM \citep{2009COFKM} is a multi-view fuzzy clustering approach developed based on FCM. To handle multi-view data, CoFKM combines the first and third strategies. For the first strategy, the term of the average disagreement between any pairs of the views $\frac{1}{P-1}(\sum\limits_{p\neq p'}\sum\limits_{i=1}^{N}\sum\limits_{c=1}^{K}((u_{ci}^{p'})^m-(u_{ci}^{p})^m)(d_{ic}^{p})^2)$ is integrated into the objective function. By minimizing the summation of the standard objective function of FCM of each view and the pairwise disagreement term, the membership $u_{ci}^{p}$ of each view is achieved. Then the third strategy is applied in which the final consensus fuzzy membership $u_{ci}^{\ast}$ is generated by calculating the geometric mean of membership of all views as follows:
\begin{equation}
u_{ci}^{\ast}=\sqrt[P]{\prod_{p=1}^{P}u_{ci}^{p}}
\label{equation:ucistar}
\end{equation}
A parameter $\eta$ is used in the objective function to control the weight of the disagreement term, however in CoFKM each view is treated equally and the weight of each view is not considered. As described in \citep{2014WVCOFCM}, it may degrade the clustering performance in  scenarios where some views are noisy and not reliable.
\subsection{WV-Co-FCM}
In \citep{2014WVCOFCM}, based on similar strategies as applied in Co-FKM, WV-Co-FCM is proposed to handle multi-view data. Same as Co-FKM, the fuzzy membership for each object in each view is first calculated in WV-Co-FCM. Then an additional step is needed to calculate the final consensus membership. There are mainly three differences between the two approaches. First, instead of using standard FCM, WV-Co-FCM is based on GIFP-FCM \citep{zhu2009generalized} in which the term $\sum\limits_{c=1}^{K}(u_{ci}(1-u_{ci}^{m-1}))$  is added to enhance the fuzzy membership. Second, the weight for each view is considered in WV-Co-FCM and the entropy regularization term of the weight is introduced into the objective function. Third, instead of using the geometric mean in Co-FKM, the final consensus membership $u_{ci}^{\ast}$ is generated based on the weight of each view as follows:
\begin{equation}
u_{ci}^{\ast}=\sum\limits_{p=1}^{P}w_{p}u_{ci}^{p}
\label{equation:ucistar2}
\end{equation}

As discussed above, both Co-FKM and WV-Co-FCM use an additional step to achieve the final consensus clustering results. For Co-FKM, as discussed above, the weight of each view is not considered which may degrade the clustering accuracy. For WV-Co-FCM, the weights are only used in the final step. In other words, the membership of each object and the cluster centroids of each view are updated independently without considering the influence of the weights.
In our method, similar to the strategy used in RMKMC, minimaxMVSC and MultiNMF, we formulate the final consensus membership directly into the objective function. Moreover, inspired by minimaxMVSC, the minimax optimization instead of direct minimization of the objective function is used in our approach. The maximum of the weighted summation of the objective function of each view is minimized. In other words, the view with larger cost measured by MinimaxFCM objective function will be given a higher weight so the cost from the view will be suppressed more vigorously/robustly than that from other views.  Hence better consensus results can be achieved. The appropriate consensus membership and the weight of each view can be obtained simultaneously in the proposed MinimaxFCM clustering process. The consensus membership and the weight of each view is achieved simultaneously in the clustering process. Next, we present our new multi-view fuzzy clustering approach called MinimaxFCM, including the detailed formulation, derivation and an in-depth analysis.
\section{The proposed approach}
In this section, first the objective function of the proposed approach MinimaxFCM is formulated. The updating rules are derived by applying the Lagrangian Multiplier method. Next, we introduce the algorithm of MinimaxFCM including detailed steps. The time complexity of the algorithm will be discussed as well.
\subsection{Fomulation of MinimaxFCM}
We formulate the multi-view fuzzy clustering as a minimax optimization as follows:
\begin{equation}
 \min\limits_{U^{\ast}, \{V^{(p)}\}_{p=1}^{P}}\quad\max\limits_{ \{\alpha^{(p)}\}_{p=1}^{P}}\;\sum\limits_{p=1}^{P}(\alpha^{(p)})^{\gamma}Q^{(p)}
\label{equation:JMVFC}
\end{equation}
here,
\begin{equation}
Q^{(p)} = \sum\limits_{c=1}^{K}\sum\limits_{i=1}^{N}(u_{ci}^{\ast})^{m}\|x_{i}^{(p)}-v_{c}^{(p)}\|^2
\label{equation:QP}
\end{equation}
subject to
\begin{equation}
\sum\limits_{c=1}^{K}u_{ci}^{\ast}= 1, \text{\,for\,}  i = 1,2,...,N
\label{equation:uci}
\end{equation}
\begin{align}
u_{ci}^{\ast}\geq 0, \text{\,for\,} c = 1,2,...,K,  i = 1,2,...N
\label{equation:uci2}
\end{align}
\begin{equation}
\sum\limits_{p=1}^{P}\alpha^{(p)}= 1
\label{equation:alphap}
\end{equation}
\begin{align}
\alpha^{(p)}\geq 0, \text{\,for\,} p = 1,2,...,P
\label{equation:alphap2}
\end{align}
 In the formulation,
 $U^{\ast}$ is the $K\times N$ membership matrix whose element in row $c$ and column $i$ is $u_{ci}^{\ast}$. $V^{(p)}$ is the $D^{(p)}\times K$ centroid matrix of $p_{th}$ view where the $c_{th}$ column is the centroid of cluster $c$ in $p_{th}$ view. Here $D^{(p)}$ is the dimension of the objects in $p_{th}$ view.
 $Q^{(p)}$ can be considered as the cost of  $p_{th}$ view which is the standard objective function of Fuzzy c means(FCM). $(\alpha^{(p)})^{\gamma}$ is the weight of $p_{th}$ view.  The parameter $\gamma \in [0, 1)$ controls the distribution of weights $(\alpha^{(p)})^{\gamma}$ for different views. $m>1$ is the fuzzifier for fuzzy clustering which controls the fuzziness of the membership.

 The clustering goal is to conduct a minimax optimization  on the objective function $\sum\limits_{p=1}^{P}(\alpha^{(p)})^{\gamma}Q^{(p)}$, and subject to the constraints in  (\ref{equation:uci}), (\ref{equation:uci2}), (\ref{equation:alphap}) and (\ref{equation:alphap2}). In this new minimax formulation for multi-view fuzzy clustering, the consensus clustering result integrating heterogeneous views of data is generated directly based on the consensus membership  $u_{ci}^{\ast}$. In addition, the weights for each view are automatically determined based on minimax optimization, without specifying the weights by users. Moreover, by using  minimax optimization, the different views are integrated harmonically by weighting each cost term $Q^{(p)}$ differently.
 \subsection{Optimization}
 It is difficult to solve the variables $u_{ci}^{\ast}$, $v_{c}^{(p)}$ and $\alpha^{(p)}$ in (\ref{equation:JMVFC}) directly because (\ref{equation:JMVFC}) is nonconvex. However, as we observed that the objective function is convex w.r.t $u_{ci}^{\ast}$ and $v_{c}^{(p)}$ and is concave w.r.t  $\alpha^{(p)}$, therefore, similar to FCM, alternating optimizaiton (AO) can be used to solve the optimization problem by solving one variable with others fixed.
  \subsubsection{Minimization: Fixing $v_{c}^{(p)}$, $\alpha^{(p)}$and updating $u_{ci}^{\ast}$}
  The Lagrangian Multiplier method is applied to solve the optimization problem of the objective function with constraints. The Lagrangian function considering the constraints is given as follows:
 \begin{equation}
\begin{split}
L_{MinimaxFCM}
 &=J_{MinimaxFCM} \\
 &+ \sum\limits_{i=1}^{N}\lambda_{i}(\sum\limits_{c=1}^{K}u_{ci}^{\ast}-1)
  + \beta(\sum\limits_{p=1}^{P}\alpha^{(p)}-1) \\
\end{split}
\label{equation:LSSPFC}
\end{equation}
where the $J_{MinimaxFCM}$ represents the objective function of MinimaxFCM $\sum\limits_{p=1}^{P}(\alpha^{(p)})^{\gamma}Q^{(p)}$. $\lambda_{i}$ and $\beta$ are the Lagrange multipliers. The condition for solving $u_{ci}^{\ast}$ is as follows:
\begin{equation}
\frac{\partial_{L_{MinimaxFCM}}}{\partial_{u_{ci}^{\ast}}} = 0
\label{equation:UCIAST}
\end{equation}
Based on (\ref{equation:UCIAST}) and constraint  (\ref{equation:uci}),  the updating rule of $u_{ci}^{\ast}$ can be derived as follow:
\begin{equation}
u_{ci}^{\ast} = \left[\sum\limits_{j=1}^{K}\left(\frac{\sum\limits_{p=1}^{P}(\alpha^{(p)})^{\gamma}\|x_{i}^{(p)}-v_{c}^{(p)}\|^{2}}{\sum\limits_{p=1}^{P}(\alpha^{(p)})^{\gamma}\|x_{i}^{(p)}-v_{j}^{(p)}\|^{2}}\right)^{\frac{1}{m-1}}\right]^{-1}
\label{equation:UCIAST1}
\end{equation}
As shown in (\ref{equation:UCIAST1}), the weight $(\alpha^{(p)})^{\gamma}$ for each view is considered in the updating for $u_{ci}^{\ast}$.
\subsubsection{Minimization: Fixing $u_{ci}^{\ast}$, $\alpha^{(p)}$and updating $v_{c}^{(p)}$}
By taking derivative of $L_{MinimaxFCM}$ with respect to $v_{c}^{(p)}$, we get:
\begin{equation}
\frac{\partial_{L_{MinimaxFCM}}}{\partial_{v_{c}^{(p)}}} = -2\sum\limits_{i=1}^{N}(\alpha^{(p)})^{\gamma}(u_{ci}^{\ast})^{m}(x_{i}^{(p)}-v_{c}^{(p)})
\label{equation:VCP}
\end{equation}
The updating rule of $v_{c}^{(p)}$ is derived as follow by setting  (\ref{equation:VCP}) to be 0:
\begin{equation}
v_{c}^{(p)} = \frac{\sum\limits_{i=1}^{N}(u_{ci}^{\ast})^{m}x_{i}^{(p)}}{\sum\limits_{i=1}^{N}(u_{ci}^{\ast})^{m}}
\label{equation:VCP1}
\end{equation}
As shown in (\ref{equation:VCP1}), the updating of the centroids of each view is the same as standard FCM.
\subsubsection{Maximization: Fixing $u_{ci}^{\ast}$, $v_{c}^{(p)}$and updating $\alpha^{(p)}$}
Based on the Lagrangian Multiplier method, the condition for solving $\alpha^{(p)}$ is as follows:
\begin{equation}
\frac{\partial_{L_{MinimaxFCM}}}{\partial_{\alpha^{(p)}}} = 0
\label{equation:ALPHAP}
\end{equation}
Based on (\ref{equation:ALPHAP}) and constraint  (\ref{equation:alphap}), the updating rule $\alpha^{(p)}$ is given as follows:
\begin{equation}
\alpha^{(p)} = \left[\sum\limits_{j=1}^{P}\left(\frac{Q^{(p)}}{Q^{(j)}}\right)^{\frac{1}{\gamma-1}}\right]^{-1}
\label{equation:ALPHAP1}
\end{equation}
Here the cost term $Q^{(p)}$ is the weighted distance summation of all the data points under the $p_{th}$ view to its corresponding centroid. The larger the value of $Q^{(p)}$ is, the larger cost this view will contribute to the objective function. From (\ref{equation:ALPHAP1}), we can see that the larger the cost of the $p_{th}$ view is, the higher value that will be assigned to $\alpha^{(p)}$ which leads to the maximum of the weighted cost. The maximum is minimized with respect to the membership and centroids in order to suppress the high cost views and achieve a harmonic consensus clustering result.  Next, we propose the details of the multi-view fuzzy clustering algorithm based on the minimax optimization.
\subsection{MinimaxFCM Algorithm}
The details of the algorithm of the proposed MinimaxFCM approach are outlined in Algorithm 1 as follows. First, the data set $X$ is represented by multiple views $\{X^{(1)}, ... X^{(P)}\}$ calculated from different features respectively. The centroids $V^{(p)}$ and $\alpha^{(p)}$ of each view are initialized. $\alpha^{(p)}$ is initialized as $\frac{1}{P}$ to make the weight be uniform for each view. Then, the consensus membership $u_{ci}^{\ast}$, the centroids $v_{c}^{(p)}$ for each view,  and $\alpha^{(p)}$  for each view are updated by using (\ref{equation:UCIAST1}), (\ref{equation:VCP1}) and (\ref{equation:ALPHAP}) respectively. Step 4-16 are repeated until the convergence condition is satisfied. In the final step, the cluster indicator $\textbf{q}$ is determined for each object. $\textbf{q}_{j}$ is the cluster number which object $j$ belongs to. This is achieved by assigning object $j$ to the cluster $c$ which has the largest consensus membership $u_{cj}^{\ast}, c\in \{1, 2, ... K\}$.
\begin{center}
\begin{tabular}{ll}
\toprule
\textbf{Algorithm 1:} MinimaxFCM\\\hline
\textbf{Input:} Data set of $P$ views $\{X^{(1)}, ... X^{(P)}\}$ with size $N$ \\
\quad\quad\quad Cluster Number $K$, stopping criterion $\epsilon$, fuzzifier $m$\\
\quad\quad\quad Parameter $\gamma$\\
\textbf{Output:}  Cluster Indicator $\textbf{q}$\\
\quad\quad\quad\; Cluster centroids $V^{(p)}$ for each view\\
\quad\quad\quad\; The weight $(\alpha^{(p)})^{\gamma}$ for each view\\
\textbf{Method:} \\
\quad {\scriptsize 1} Initialize centroids $V^{(p)}$ for each view. \\
\quad {\scriptsize 2} Initialize $\alpha^{(p)}=\frac{1}{P}$ for each view\\
\quad {\scriptsize 3} Set $t=0$\\
\quad\,\,    \textbf{Repeat} \\
\quad {\scriptsize 4}\quad     \textbf{for} $c=1 \,\text{to} \,K$ \\
\quad {\scriptsize 5}\quad\,\,    \textbf{for} $i=1 \,\text{to} \,N$ \\
\quad {\scriptsize 6}\quad\quad    Update $u_{ci}^{\ast}$ using equation (\ref{equation:UCIAST1});  \\
\quad {\scriptsize 7}\quad\,\,     \textbf{end for}\\
\quad {\scriptsize 8}\quad     \textbf{end for}\\
\quad {\scriptsize 9}\quad     \textbf{for} $c=1 \,\text{to} \,K$ \\
\quad {\scriptsize 10}\quad\,\,    \textbf{for} $p=1 \,\text{to} \,P$ \\
\quad {\scriptsize 11}\quad\,\,    Update $v_{c}^{(p)}$ using equation (\ref{equation:VCP1}); \\
\quad {\scriptsize 12}\quad\,\,     \textbf{end for}\\
\quad {\scriptsize 13}\quad     \textbf{end for}\\
\quad {\scriptsize 14}\quad     \textbf{for} $p=1 \,\text{to} \,P$ \\
\quad {\scriptsize 15}\quad\quad  Update $\alpha^{(p)}$ using equation (\ref{equation:ALPHAP}); \\
\quad {\scriptsize 16}\quad     \textbf{end for}\\
\quad {\scriptsize 17}\quad    Update $t=t+1$\\
\quad\,\,    \textbf{Until} ($\parallel (U^{\ast})^{t+1} - (U^{\ast})^{t} \parallel < \epsilon$)\\
\quad {\scriptsize 18}\quad     $\textbf{q}_{j} = arg\,max_{1\leq c \leq K}u_{cj}^{\ast}, j = 1,2,...N.$\,\,\\
\hline
\end{tabular}
\end{center}
There are two parameters including the fuzzifier $m$ and parameter $\gamma$ need to be set before running the algorithm.
The parameter $\gamma$ controls the distribution of weight $(\alpha^{(p)})^{\gamma}$ for different views. It can be seen from (\ref{equation:ALPHAP1}) that when $\gamma\rightarrow 0$, the weight $(\alpha^{(p)})^{\gamma}$ of each view will become close to each other. When $\gamma\rightarrow 1$, the weight of view $p$ whose cost term $Q_{(p)}$ is the largest among all views will be assigned as 1, and the other views will be assigned as 0.

The time complexity of MinimaxFCM is $O(T\cdot(3\cdot P\cdot N\cdot K))$ considering the number of iterations $T$, number of objects $N$, number of clusters $K$ and number of views $P$.  $O(P\cdot N\cdot K)$ is the time complexity for updating $u_{ci}^{\ast}$, $v_{c}^{(p)}$,  and $\alpha^{(p)}$, respectively. Note that different from the graph based multi-view algorithms such as \citep{2011NIPScosc} and \citep{wang2014multi}, the time complexity of our multi-view fuzzy clustering is similar to FCM. The graph construction and eigen decomposition whose time complexity are $O(N^{2})$ and $O(N^{3})$ respectively in graph based algorithms are time consuming.
\section{Experimental results}
In this section, experimental studies of the proposed approach are conducted on different kinds of multi-view data sets including image and document data sets.
In the experiments, we compare the performace of MinimaxFCM with six related approaches on multi-view data clustering.
The experiments implemented in Matlab were conducted on a PC with four cores of Intel I5-2400 with 8 gigabytes of memory.
\subsection{Data sets}
Nine data sets as summarized in Table.~\ref{table:datainfo}  were used for the experimental study and comparisons.

\begin{table}[ht]
\footnotesize
\caption{The characteristics of the multi-view data sets} 
\centering 
\begin{tabular}{c c c c c} 
\hline
Data Sets & No. of views & No. of classes & No. of objects & No. of dimension\\ [0.5ex] 
\hline 
Multiple features & 6 & 10 & 2000 & 649 \\ 
Image segmentation & 2 & 7& 2310 & 19 \\
Corel1 & 7 & 4& 400 & 338 \\
Corel2 & 7 & 4& 400 & 338 \\
Corel3 & 7 & 4& 400 & 338 \\
Corel4 & 7 & 4& 400 & 338 \\
Corel5 & 7 & 4& 400 & 338 \\
3-Sources document & 3 & 6& 169 & 10259 \\
Reuters multilingual data & 5 & 6& 1500 & 107783 \\[1ex] 
\hline 
\end{tabular}
\label{table:datainfo} 
\end{table}

Multiple features (MF)\footnote{This data set can be downloaded on https://archive.ics.uci.edu/ml/datasets/Multiple+Features.}: This data set consists of 2000 handwritten digit images (0-9) extracted from a collection of Dutch utility maps. It has 10 classes and each class has 200 images. Each object is described by 6 different views (Fourier coefficients, profile correlations, Karhunen-Love coefficients, pixel averages, Zernike moments, morphological features).

Image segmentation (IS) data set\footnote{This data set can be downloaded on https://archive.ics.uci.edu/ml/datasets/Image+Segmentation.}: This data set is composed of 2310 outdoor images which have 7 classes. Each image is represented by 19 features. The features can be considered as two views which are shape view and RGB view. The shape view consists of 9 features which describe the shape information of each image. The RGB view consists of 10 features which describe the RGB values of each image.

\begin{figure}
\centering
\includegraphics[width=3in]{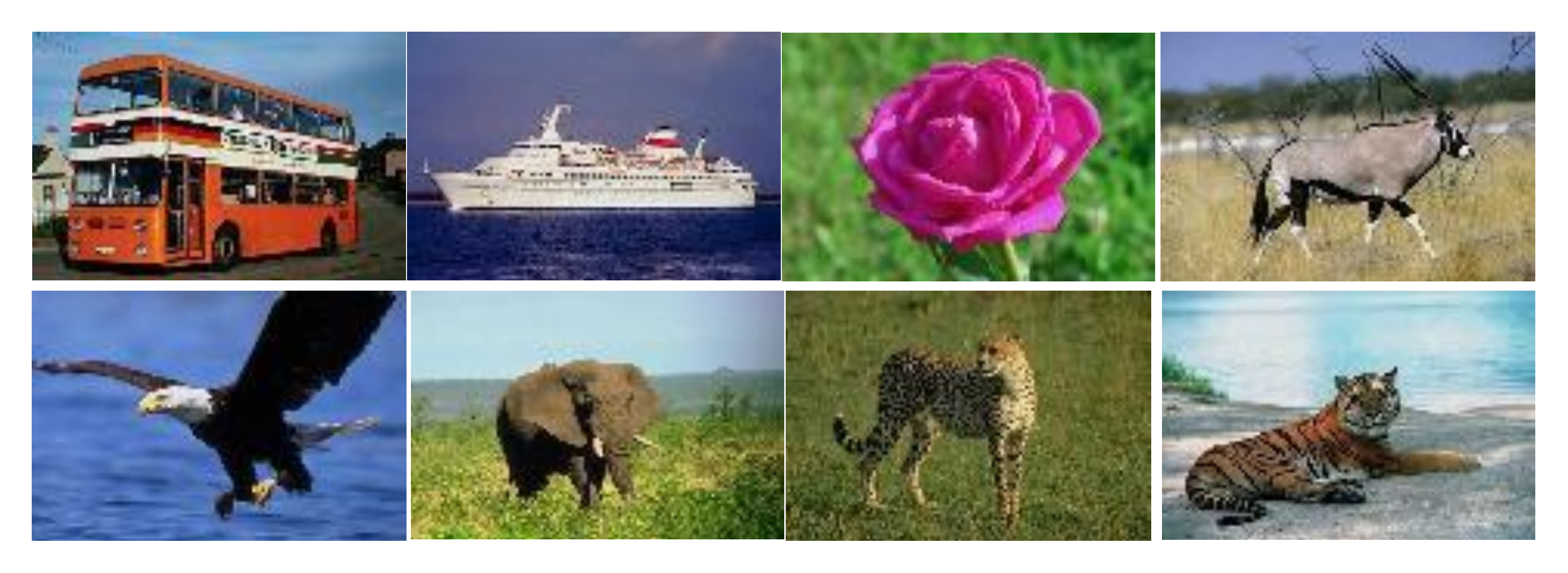}
\caption{Image examples in Corel data set}
\label{fig:corelSamples}
\end{figure}

Corel Image data set \footnote{This data set can be downloaded on http://www.cs.virginia.edu/~xj3a/research/CBIR/Download.htm.}: This data set is a part of the popular corel image data set which consists of  34 classes. Each class has 100 images. Some of the image examples are shown in Fig.~\ref{fig:corelSamples}. Each image is represented by 7 different views including three color-related views and four texture-related views. Table.~\ref{table:corelInfo} shows details of the 7 views. We extracted several four class subsets and tested the representative ones as shown in Table.~\ref{table:corelsubset}.

\begin{table}[]
\caption{The 7 views of the Corel Image data set} 
\centering 
\begin{tabular}{c| c |c} 
\hline
View Categories & View name & Dimension\\ [0.5ex] 
\hline 
\multirow{3}{*}{Color Views} & Color Histogram & 64  \\ 
 & Color Moment & 9  \\
 & Color Coherence & 128  \\
 \hline
\multirow{4}{*}{Texture Views} & Coarseness of Tamura Texture& 10  \\
 & Directionality of  Tamura Texture& 8  \\
 & Wavelet Texture & 104  \\
 & MASAR Texture & 15 \\[1ex] 
\hline 
\end{tabular}
\label{table:corelInfo} 
\end{table}

\begin{table}[]
\caption{Classes contained in the tested Corel subsets} 
\centering 
\begin{tabular}{c| c c c c} 
\hline
Subsets & \multicolumn{4}{c}{Classes} \\ [1ex] 
\hline 
Corel1& Owls  &Tigers  &Trains  &Ships   \\ 
Corel2& Buses  &Leopards &Trains &Ships   \\
Corel3 & Buses  &Leopards  &Cars  &Deer  \\
Corel4 & Owls  &Tigers  &Eagles  &Flowers  \\
Corel5 & Eagles  &Elephants  &Cars & Deer   \\[1ex] 
\hline 
\end{tabular}
\label{table:corelsubset} 
\end{table}
3-Sources document data set\footnote{This data set can be downloaded on http://mlg.ucd.ie/datasets/3sources.html.}: This data set consists of  948 news articles covering 416 distinct news stories. They are collected from three online news sources: BBC, Guardian and Reuters. We selected 169 news articles which are reported in all three sources.  It has 6 topic classes which are business, entertainment, health, politics, sport and technique. Each article is described by 3 views which are the three sources.

Reuters multilingual data set: This data set contains documents originally written in five different languages (English, French, German, Spanish and Italian) and their translations \citep{amini2009learning}. This multilingual data set covers a common set of six classes. We use documents originally in English as the first view and their four translations as the other four views. We randomly sample 1500 documents from this collection with each of the 6 classes having 250 documents.

\subsection{Experimental settings}
We firstly compare the performance of the proposed MinimaxFCM with its corresponding single-view counterpart. In addition, we compare the results of our method with those of the baseline method naive multi-view fuzzy clustering which is implemented by simply using the concatenated features of all views as input to the FCM clustering algorithm. In order to demonstrate the effectiveness of MinimaxFCM, different kinds of multi-view clustering approaches are also comprehensively compared with. This consists of two fuzzy clustering based approaches including Co-FKM \citep{2009COFKM} and WV-Co-FCM \citep{2014WVCOFCM}; K means based RMKMC, nonnegative matrix factorization based MultiNMF; spectral clustering based approach using minimax optimization MinimaxMVSC. The six compared approaches and their parameter settings are summarized as follows:
 \begin{enumerate}
    \item \textit{FCM on Single View}: We apply standard FCM on each single view of the data sets and report the worst and best clustering results among different views.
    \item \textit{FCM on Concatenated View}: We first concatenate the features of all views and then apply standard FCM on the concatenated data.
    \item \textit{Multiview Fuzzy Clustering}: As in WV-Co-FCM \citep{2014WVCOFCM}, the grid search strategy is adopted to find the better parameters.  For Co-FKM, as recommended in \citep{2009COFKM}, the parameter $\eta$ is searched from $0\leq \eta \leq \frac{\mid P-1\mid}{\mid P\mid}$ with the step 0.01. Here $P$ is the number of views. For WV-Co-FCM, the searching method of parameters are the same as describe in \citep{2014WVCOFCM}. And we select the first updating equation (case (a) in \citep{2014WVCOFCM}) for WV-Co-FCM as the results of the four updating equations are very similar.
     \item \textit{Multiview K means Clustering}: The robust multiview k means clustering (RMKMC) \citep{2013IJCAIRmultiviewKmeans} is compared here. As recommended in \citep{2013IJCAIRmultiviewKmeans} , the parameter $\alpha$ is searched based on the logarithm of the parameter, which is, $log_{10}^{\alpha}$ in the range of [0.1 2] with step 0.2.
     \item \textit{Multiview Nonnegative Matrix Factorization}: The multiview clustering based joint Nonnegative Matrix Factorization (MultiNMF)\citep{liu2013multi} is compared. As recommended in \citep{liu2013multi}, the parameter $\lambda$ is set to be 0.01.
     \item \textit{Multiview Spectral Clustering}: The multiview spectral clustering based minimax optimization (MinimaxMVSC) \citep{wang2014multi} is compared. The parameter $\gamma$ is searched in the range of [0.1 0.9] with step 0.1.
\end{enumerate}
The parameter setting for MinimaxFCM is similar to that in \citep{2009COFKM}. The parameter $\gamma$ is searched from  [0.1 0.9]  with the step 0.1.  For all fuzzy clustering based approaches, the fuzzifier $m$ is set by searching from [1.1 2] with the step 0.05. The results reported are the value with the best searched parameter for each approach.

\subsection{Evaluation criterion}
Three popular external criteria \emph{Accuracy} \citep{mei2012fuzzy}, \emph{F-measure} \citep{larsen1999fast}, and \emph{Normalized Mutual Information}(NMI) \citep{strehl2003cluster} are used to evaluate the clustering results, which measure the agreement of the clustering results produced by an algorithm and the ground truth. If we refer to \textit{class} as the ground truth, and \textit{cluster} as the results of a clustering algorithm, the NMI is calculated as follows:
 \begin{equation}
 NMI = \frac{\sum\limits_{c=1}^{k}\sum\limits_{p=1}^{m}n_{c}^{p}log(\frac{n\cdot n_{c}^{p}}{n_{c}\cdot n_{p}})}{\sqrt{(\sum\limits_{c=1}^{k}n_{c}log(\frac{n_{c}}{n}))(\sum\limits_{p=1}^{m}n_{p}log(\frac{n_{p}}{n}))}} \\
 \label{equation:NMIcalc}
 \end{equation}
 where $n$ is the total number of objects, $n_{c}$ and $n_{p}$ are the numbers of objects in the $c_{th}$ \textit{cluster} and the $p_{th}$ \textit{class}, respectively, and $n_{c}^{p}$ is the number of common objects in \textit{class} $p$ and \textit{cluster} $c$.  For F-measure, the calculation based on precision and recall is as follows:
  \begin{equation}
 F-measure = \frac{2\cdot precision\cdot recall}{precision + recall} \\
 \label{equation:F-meaure}
 \end{equation}
 where,
  \begin{equation}
   precision = \frac{n_{c}^{p}}{n_{c}}\\
 \label{equation:precision}
 \end{equation}
 \begin{equation}
   recall = \frac{n_{c}^{p}}{n_{p}}\\
 \label{equation:recall}
 \end{equation}
 Accuracy is calculated as follows after obtaining a one-to-one match between \textit{clusters} and
\textit{classes}:
  \begin{equation}
 Accuracy = \sum\limits_{c=1}^{k}\frac{n_{c}^{j}}{n} \\
 \label{equation:Accuracy}
 \end{equation}
 where $n_{c}^{j}$ is the number of common objects in the $c_{th}$ cluster
and its matched class $j$. The higher the values of the three criterions are, the better the clustering result is. The value is equal to 1 only when the clustering result is same as the ground truth.

\subsection{Initialization}
To make MinimaxFCM be more robust to initialization, we initialized the K centroids for each view based on the method used in \citep{krishnapuram2001low}.  K objects in each view are selected as the initial K centroids  for each view. For each view, we select the object which has the minimum distance to all the other objects as the first centroid. The remaining centroids are chosen consecutively by selecting the objects that maximize their minimal distance with existing centroids.  Based on the selection mechanism, convergence to a bad local optimum may be avoided because the centroids are distributed evenly in the data space. The detailed steps of initialization of MinimaxFCM are as follows.
\begin{center}
\begin{tabular}{ll}
\hline
\textbf{Initialization for MinimaxFCM} \\\hline
Set the number of clusters $K$\\
{\scriptsize 1}     \textbf{for} $p=1 \,\text{to} \,P$ \\
{\scriptsize 2}\quad     Calculate the first centroid:\\
{\scriptsize 3}\quad \qquad $t=arg\,min_{1\leq{j}\leq{n}}\sum\limits_{i=1}^{n}d_{ij}^{p};$ \\
{\scriptsize 4}\quad \qquad $\text{\,first centroid:\,}v_{1}=x_{t}^{p}$\\
{\scriptsize 5}\quad Centroids set $V_{}^{p}=\{{v_{1}^{p}}\}, m=1$; \\
\quad\, \textbf{Repeat} \\
{\scriptsize 6}\quad \qquad $m=m+1$\\
{\scriptsize 7}\quad \qquad $t=arg\,max_{1\leq{i}\leq{n};x_{i}\not\in{V^{p}}}min_{1\leq{k}\leq{|V^{p}|}}Dis(v_{k}^{p},x_{i}^{p});$\\
{\scriptsize 8}\quad \qquad $\text{\,centroid:\,}v_{m}^{p}=x_{t}^{p}$ \\
{\scriptsize 9}\quad \qquad $V_{}^{p}=V_{}^{p}\cup\{{v_{m}^{p}}\}$;\\
\quad\, \textbf{Until}(m=K) \\
{\scriptsize 10}\,    \textbf{end for}\\
\hline
\end{tabular}
\end{center}
For fair comparison, the same initialization method is applied in standard FCM, Co-FKM and WV-Co-FCM to initialize the centroids. For RMKMC and MultiNMF,  we initialize the cluster centroid matrix which is composed of the centroids selected based on the same method. The same method is also applied in K-means which is used as the final step of MinimaxMVSC.
\subsection{Results on image data sets}
For the image data sets Multiple features (MF), Image segmentation (IS) and five subsets of the Corel data set, to get the comparable cost $Q^{(p)}$ of each view, we adopt the method used in Co-FKM \citep{2009COFKM} to normalize each view. We normalize each feature to unit variance, and apply a weight with value $\frac{1}{\sqrt{D^{(p)}}}$ on the data in $p_{th}$ view. Here $D^{(p)}$ is the dimension of $p_{th}$ view. The Euclidean distance measure is used to calculate the distance. For MultiNMF, as described in \citep{liu2013multi}, the data is preprocessed to make $|| X^{p}||=||\sum\limits_{i=1}^{N}x_{i}^{p}||_{1}=1$. For MinimaxMVSC, as described in \citep{wang2014multi}, the similarity matrix is constructed based on the Gaussian kernel in which Euclidean distance is used. Note that the above initialization method will generate the same set of initial centroids, hence the clustering results of each run is same for each approach. The accuracy, NMI and F-measure results of MF, IS and Corel data sets are shown in Table.~\ref{table:mf}, ~\ref{table:is} and ~\ref{table:imagedataresults}, respectively.
\begin{table}[!htbp]
\normalsize
\caption{Comparison of the approaches on Multiple features(MF) data set} 
\centering 
\begin{tabular}{c c c c}
\hline
 & Accuracy & NMI & F-measure \\ [0.5ex] 
\hline 
Worst Single View & $0.3935$ & $\,0.4786$ & $\,0.4583 $ \\
Best Single View & $0.7833$ & $\,0.7424$ & $\,0.7990 $ \\ 
Concatenated View & $0.8476$ & $\,0.8081$ & $\,0.8624$ \\
RMKMC & $0.8849$ & $\,0.8442$ & $\,0.8913$ \\
MultiNMF & $0.9025$ & $\,0.8308$ & $\,0.9022$ \\
Co-FKM & $0.9048$ & $\,0.8834$ & $\,0.9142$ \\
WV-Co-FCM & $0.9358$ & $\,0.8852$ & $\,0.9359$ \\
MinimaxMVSC & $0.9200$ & $\,0.8448$ & $\,0.9200$ \\
MinimaxFCM & $\textbf{0.9610}$ & $\,\textbf{0.9151}$ & $\,\textbf{0.9611}$ \\ 
\hline 
\end{tabular}
\label{table:mf}
\end{table}

\begin{table}[!htbp]
\normalsize
\caption{Comparison of the approaches on Image segmentation(IS) data set}
\centering
\begin{tabular}{c c c c}
\hline
 & Accuracy & NMI & F-measure \\ [0.5ex] 
\hline 
Worst Single View & $0.2815$ & $\,0.1420$ & $\,0.3083 $ \\
Best Single View & $0.5335$ & $\,0.5601$ & $\,0.5918$ \\ 
Concatenated View & $0.4168$ & $\,0.3680$ & $\,0.4412$ \\
RMKMC & $0.6319$ & $\,0.6128$ & $\,0.6364$ \\
MultiNMF & $0.6173$ & $\,0.5799$ & $\,0.6449$ \\
Co-FKM & $0.6334$ & $\,0.5923$ & $\,0.6214$ \\
WV-Co-FCM & $0.6338$ & $\,0.6139$ & $\,0.6614$ \\
MinimaxMVSC & $0.6658$ & $\,0.6010$ & $\,0.6653$ \\
MinimaxFCM & $\textbf{0.6723}$ & $\,\textbf{0.6344}$ & $\,\textbf{0.6988}$ \\ 
\hline 
\end{tabular}
\label{table:is}
\end{table}

\begin{table}[!htbp]
\normalsize
\caption{Comparison of the approaches on Corel data set}
\label{table:imagedataresults} 
\centering
\subfloat[][Corel1]{
\begin{tabular}{c c c c}
\hline
 & Accuracy & NMI & F-measure \\ [0.5ex] 
\hline 
Worst Single View & $0.4025$ & $\,0.1608$ & $\,0.4795 $ \\
Best Single View & $0.6950$ & $\,0.4554$ & $\,0.6988$ \\ 
Concatenated View & $0.6951$ & $\,0.4694$ & $\,0.6834$ \\
RMKMC & $0.7030$ & $\,0.5579$ & $\,0.7358$ \\
MultiNMF & $0.7800$ & $\,0.5741$ & $\,0.7806$ \\
Co-FKM & $0.7450$ & $\,0.4783$ & $\,0.7395$ \\
WV-Co-FCM & $0.7575$ & $\,0.5442$ & $\,0.7560$ \\
MinimaxMVSC & $0.7425$ & $\,0.5565$ & $\,0.7367$ \\
MinimaxFCM & $\textbf{0.8100}$ & $\,\textbf{0.6215}$ & $\,\textbf{0.8084}$ \\ 
\hline 
\end{tabular}}

\subfloat[][Corel2]{
\begin{tabular}{c c c c}
\hline
 & Accuracy & NMI & F-measure \\ [0.5ex] 
\hline 
Worst Single View & $0.4275$ & $\,0.1232$ & $\,0.4244 $ \\
Best Single View & $0.6575$ & $\,0.4205$ & $\,0.6723$ \\ 
Concatenated View & $0.6675$ & $\,0.4830$ & $\,0.6939$ \\
RMKMC & $0.6950$ & $\,0.5396$ & $\,0.7198$ \\
MultiNMF & $0.7450$ & $\,0.5282$ & $\,0.7351$ \\
Co-FKM & $0.7200$ & $\,0.5384$ & $\,0.7029$ \\
WV-Co-FCM & $0.7200$ & $\,0.5436$ & $\,0.7034$ \\
MinimaxMVSC & $0.7400$ & $\,0.5565$ & $\,0.7347$ \\
MinimaxFCM & $\textbf{0.7475}$ & $\,\textbf{0.5605}$ & $\,\textbf{0.7370}$ \\ 
\hline 
\end{tabular}}

%
\end{table}

\begin{table}
\ContinuedFloat
\normalsize
\centering
\subfloat[][Corel3]{
\begin{tabular}{c c c c}
\hline
 & Accuracy & NMI & F-measure \\ [0.5ex] 
\hline 
Worst Single View & $0.3825$ & $\,0.0677$ & $\,0.3790 $ \\
Best Single View & $0.7325$ & $\,0.4790$ & $\,0.7224$ \\ 
Concatenated View & $0.5200$ & $\,0.4205$ & $\,0.5941$ \\
RMKMC & $0.7825$ & $\,0.6008$ & $\,0.7861$ \\
MultiNMF & $0.7800$ & $\,0.5245$ & $\,0.7772$ \\
Co-FKM & $0.7725$ & $\,0.6130$ & $\,0.7700$ \\
WV-Co-FCM & $0.7800$ & $\,0.6250$ & $\,0.7785$ \\
MinimaxMVSC & $0.7825$ & $\,0.5900$ & $\,0.7825$ \\
MinimaxFCM & $\textbf{0.7975}$ & $\,\textbf{0.6378}$ & $\,\textbf{0.8005}$ \\ 
\hline 
\end{tabular}}

\subfloat[][Corel4]{
\begin{tabular}{c c c c}
\hline
 & Accuracy & NMI & F-measure \\ [0.5ex] 
\hline 
Worst Single View & $0.3450$ & $\,0.0547$ & $\,0.3590 $ \\
Best Single View & $0.7600$ & $\,0.4791$ & $\,0.7672$ \\ 
Concatenated View & $0.6150$ & $\,0.3630$ & $\,0.6400$ \\
RMKMC & $0.8625$ & $\,\textbf{0.6812}$ & $\,0.8608$ \\
MultiNMF & $0.8225$ & $\,0.5763$ & $\,0.8230$ \\
Co-FKM & $0.8550$ & $\,0.6608$ & $\,0.6534$ \\
WV-Co-FCM & $0.8575$ & $\,0.6595$ & $\,0.8564$ \\
MinimaxMVSC & $0.8575$ & $\,0.6708$ & $\,0.8557$ \\
MinimaxFCM & $\textbf{0.8650}$ & $\,0.6774$ & $\,\textbf{0.8643}$ \\ 
\hline 
\end{tabular}}
\end{table}

\begin{table}
\ContinuedFloat
\normalsize
\centering
\subfloat[][Corel5]{
\begin{tabular}{c c c c}
\hline
 & Accuracy & NMI & F-measure \\ [0.5ex] 
\hline 
Worst Single View & $0.4075$ & $\,0.0834$ & $\,0.4069 $ \\
Best Single View & $0.5900$ & $\,0.3850$ & $\,0.6172$ \\ 
Concatenated View & $0.4900$ & $\,0.3522$ & $\,0.5727$ \\
RMKMC & $0.6525$ & $\,0.4741$ & $\,0.6678$ \\
MultiNMF & $0.6550$ & $\,0.4913$ & $\,0.7039$ \\
Co-FKM & $0.6875$ & $\,0.4777$ & $\,0.6929$ \\
WV-Co-FCM & $0.6775$ & $\,0.4786$ & $\,0.6884$ \\
MinimaxMVSC & $0.6950$ & $\,0.5328$ & $\,0.7143$ \\
MinimaxFCM & $\textbf{0.7100}$ & $\,\textbf{0.5225}$ & $\,\textbf{0.7167}$ \\ 
\hline 
\end{tabular}}
\end{table}


From the tables we can see that all the multi-view clustering approaches perform better than the best single view and the concatenated view. We also observe that the concatenating method in which the features of all views are concatenated directly may not be a guarantee to generate better clustering results. For example, the results of  concatenated view of Multiple features(MF) data set are better than its best single view, while the results of  concatenated view of Image segmentation(IS) data set are worse than its best single view. The reason for this phenomenon may be that different views are not compatible with each other. MinimaxFCM is based on minimax optimization which helps to find the harmonic consensus clustering results for the data with both compatible or non-compatible views. As we can see from the tables that MinimaxFCM performs the best in almost all the data sets. Note that MinimaxFCM performs better than MinimaxMVSC in which minimax optimization is also used.

\subsection{Results on document data sets}
For document data sets (3-Sources and Reuters multilingual data), as the bag-of-words representation of documents generates features which are very sparse and high-dimensional, standard distance measures for example Euclidean distance in high dimensions are always unreliable. Therefore, for 3-Sources data, we adopt a normalization method similar to that used in \citep{liu2013multi}. For each document $x_{i}^{p}$ in the $p_{th}$ view $X_{}^{p}$, we conduct $l_{1}$ normalization such that $||x_{i}^{p}||_{1}=1$. Moreover, the cosine distance is used for the 3-Sources data set.
For Reuters multilingual data set, same as the experimental setting in \citep{2011NIPScosc}, Probabilistic Latent Semantic Analysis (PLSA) \citep{hofmann1999probabilistic} is applied to project the data to a 100-dimensional space and the clustering approaches are conducted on the low dimensional data.
Table.~\ref{table:docdataresults} shows the results of accuracy, F-measure and NMI of the two data sets. As we can see that the MinimaxFCM performs better consistently than the other approaches.
\begin{table}[!htbp]
\caption{Comparison of the approaches on two document data sets} 
\normalsize
\centering 

\subfloat[][3-Sources]{
\begin{tabular}{c c c c}
\hline
 & Accuracy & NMI & F-measure \\ [0.5ex] 
\hline 
Worst Single View & $0.5344$ & $\,0.3558 $ & $\,0.5318 $ \\
Best Single View & $0.7574$ & $\,0.6056 $ & $\,0.7464 $ \\ 
Concatenated View & $0.7041$ & $\,0.6192$ & $\,0.7066$ \\
RMKMC & $0.6331$ & $\,0.4754$ & $\,0.6063$ \\
MultiNMF & $0.5917$ & $\,0.6326$ & $\,0.6323$ \\
Co-FKM & $0.7633$ & $\,0.6516$ & $\,0.7644$ \\
WV-Co-FCM & $\textbf{0.7988}$ & $\,0.6920$ & $\,0.7934$ \\
MinimaxMVSC & $0.6450$ & $\,0.6533$ & $\,0.7106$ \\
MinimaxFCM & $0.7811$ & $\,\textbf{0.7061}$ & $\,\textbf{0.8198}$ \\ 
\hline 
\end{tabular}}


\subfloat[][Reuters multilingual]{
\begin{tabular}{c c c c}
\hline
 & Accuracy & NMI & F-measure \\ [0.5ex] 
\hline 
Worst Single View & $0.1847$ & $\,0.0034 $ & $\,0.2146 $ \\
Best Single View & $0.3467$ & $\,0.1764$ & $\,0.4007$ \\ 
Concatenated View & $0.4003$ & $\,0.2277$ & $\,0.4479$ \\
RMKMC & $0.4063 $ & $\,0.2888 $ & $\,0.4548$ \\
MultiNMF & $0.4367$ & $\,0.3022$ & $\,0.4894$ \\
Co-FKM & $0.4620 $ & $\,0.3190 $ & $\,0.4930$ \\
WV-Co-FCM & $0.4633$ & $\,0.3196$ & $\,0.5064$ \\
MinimaxMVSC & $0.4307$ & $\,0.2876$ & $\,0.4784$ \\
MinimaxFCM & $\textbf{0.5087}$ & $\,\textbf{0.3274 }$ & $\,\textbf{0.5329}$ \\ 
\hline 
\end{tabular}}
\label{table:docdataresults} 
\end{table}
\subsection{Parameter Analysis}
\begin{figure}
\centering
\begin{tabular}{ccc}
\subfloat[Multiple features]{\includegraphics[width=1.5in]{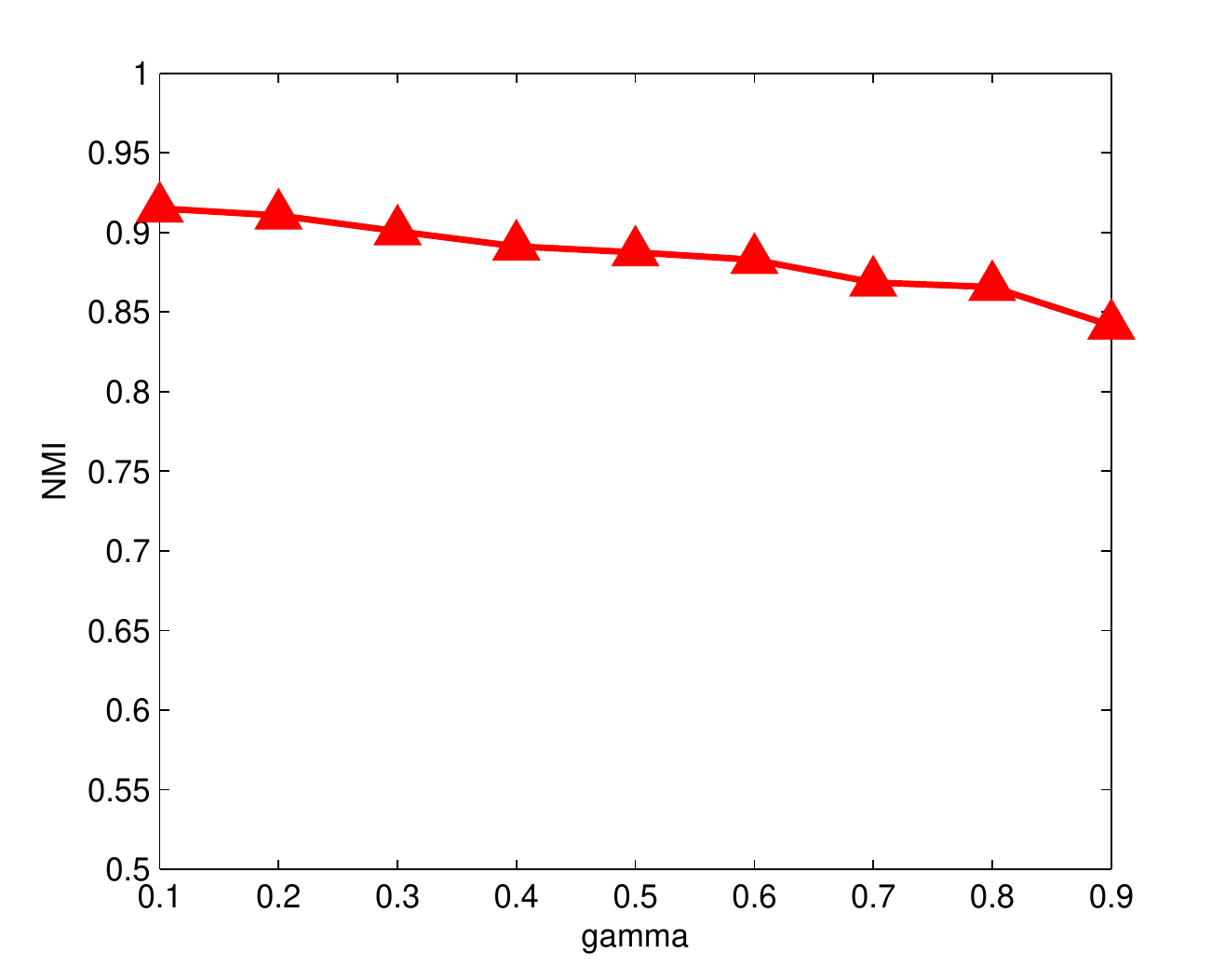}} &
\subfloat[Image segmentation]{\includegraphics[width=1.5in]{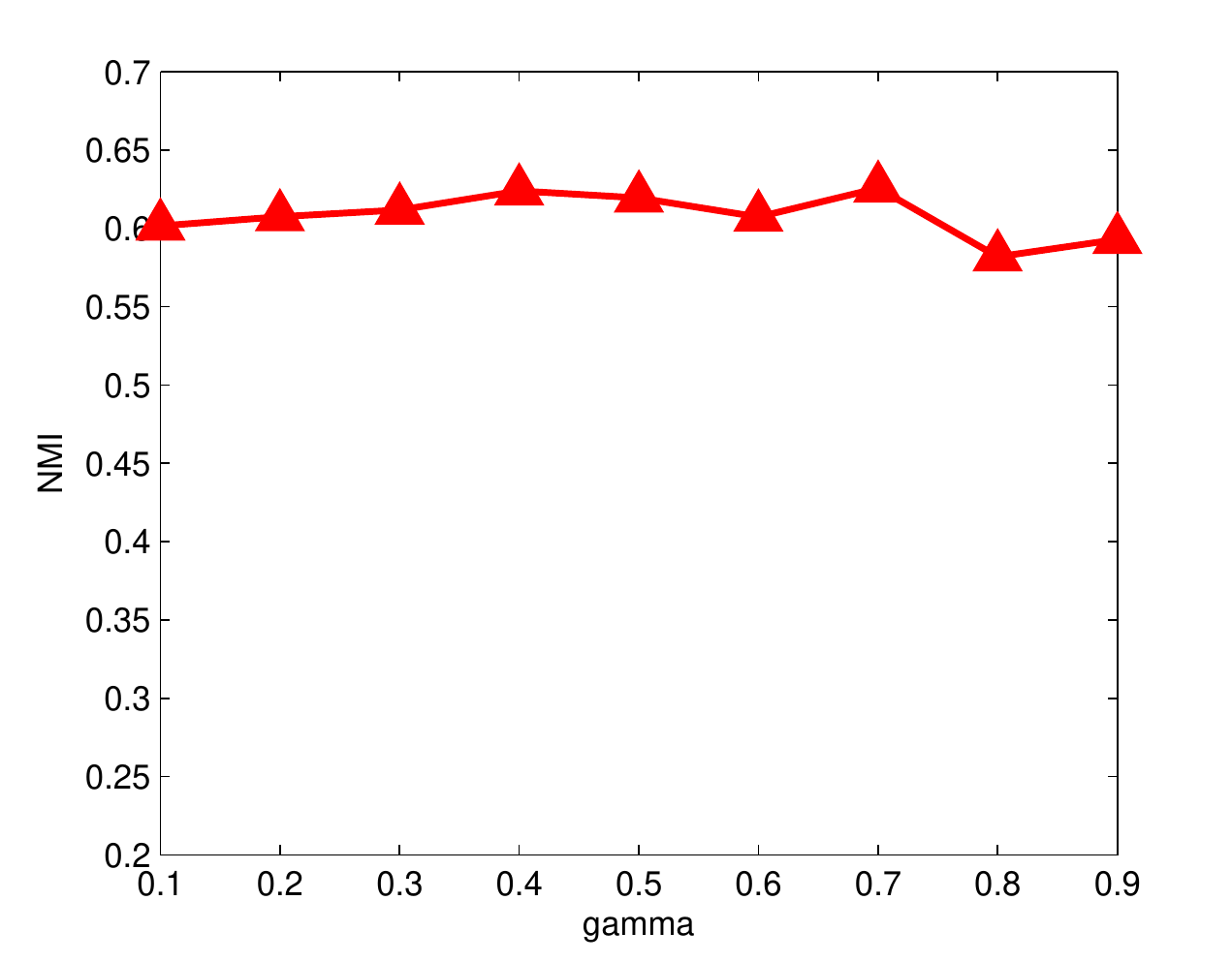}} &
\subfloat[Corel1]{\includegraphics[width=1.5in]{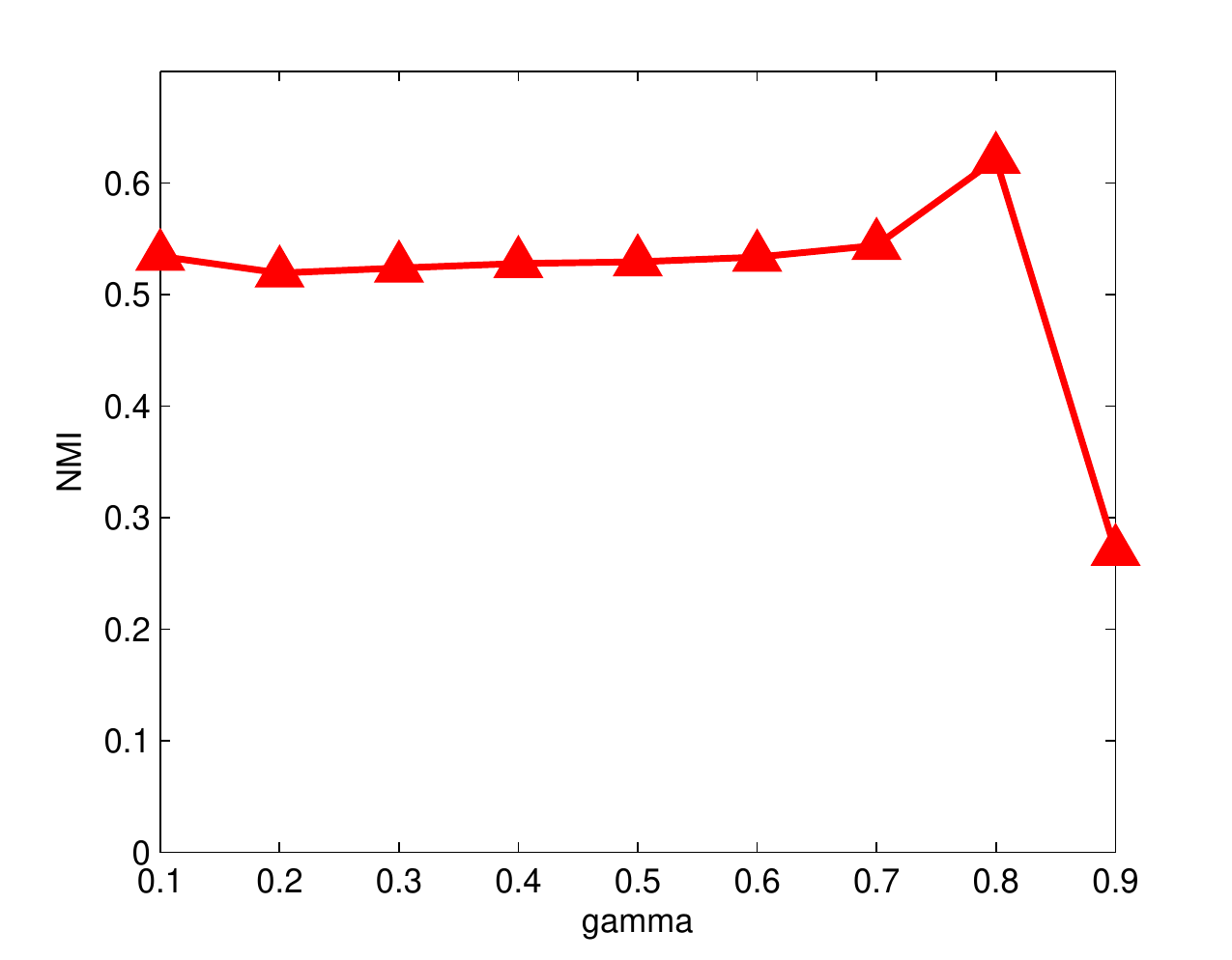}} \\
\subfloat[Corel2]{\includegraphics[width=1.5in]{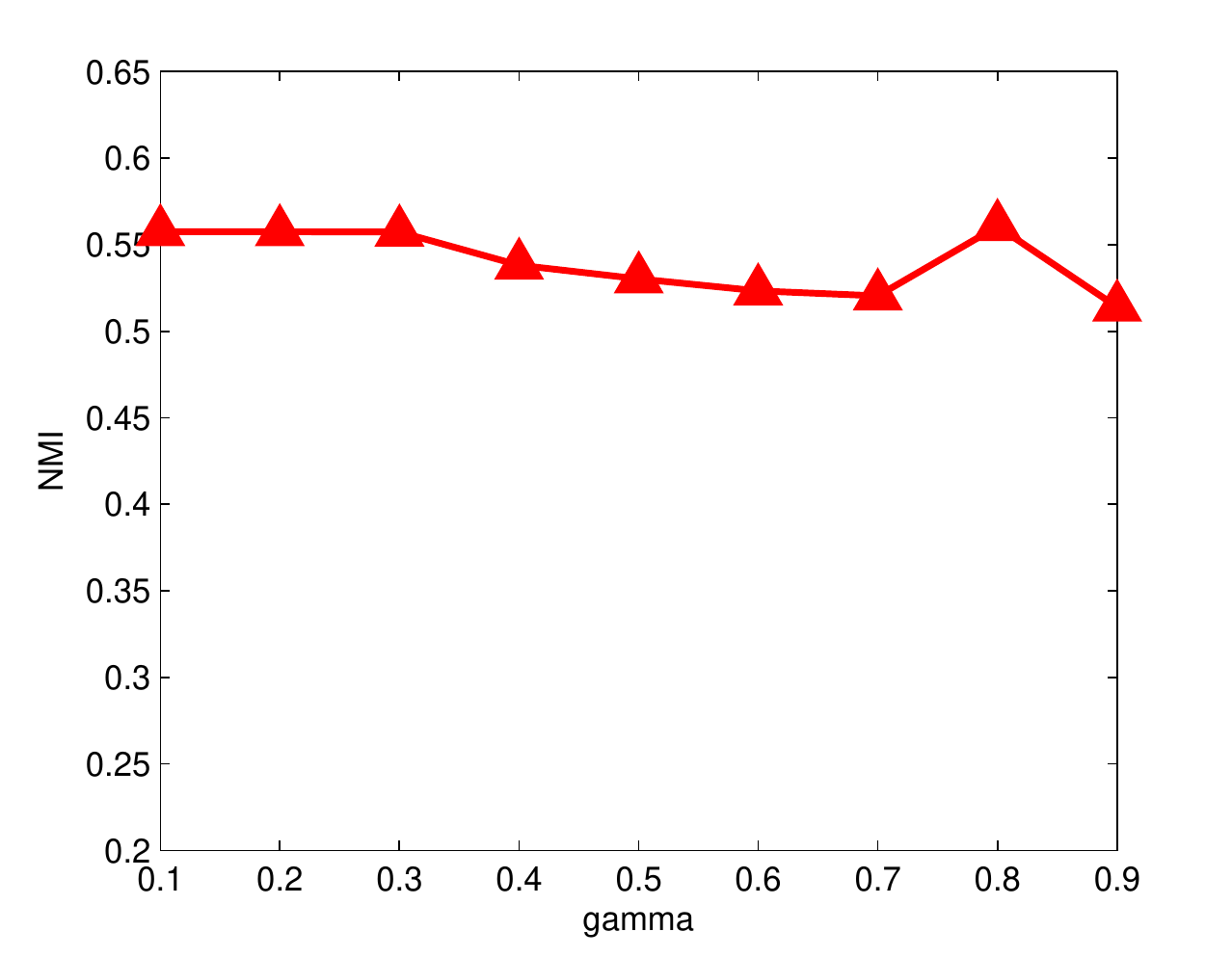}} &
\subfloat[Corel3]{\includegraphics[width=1.5in]{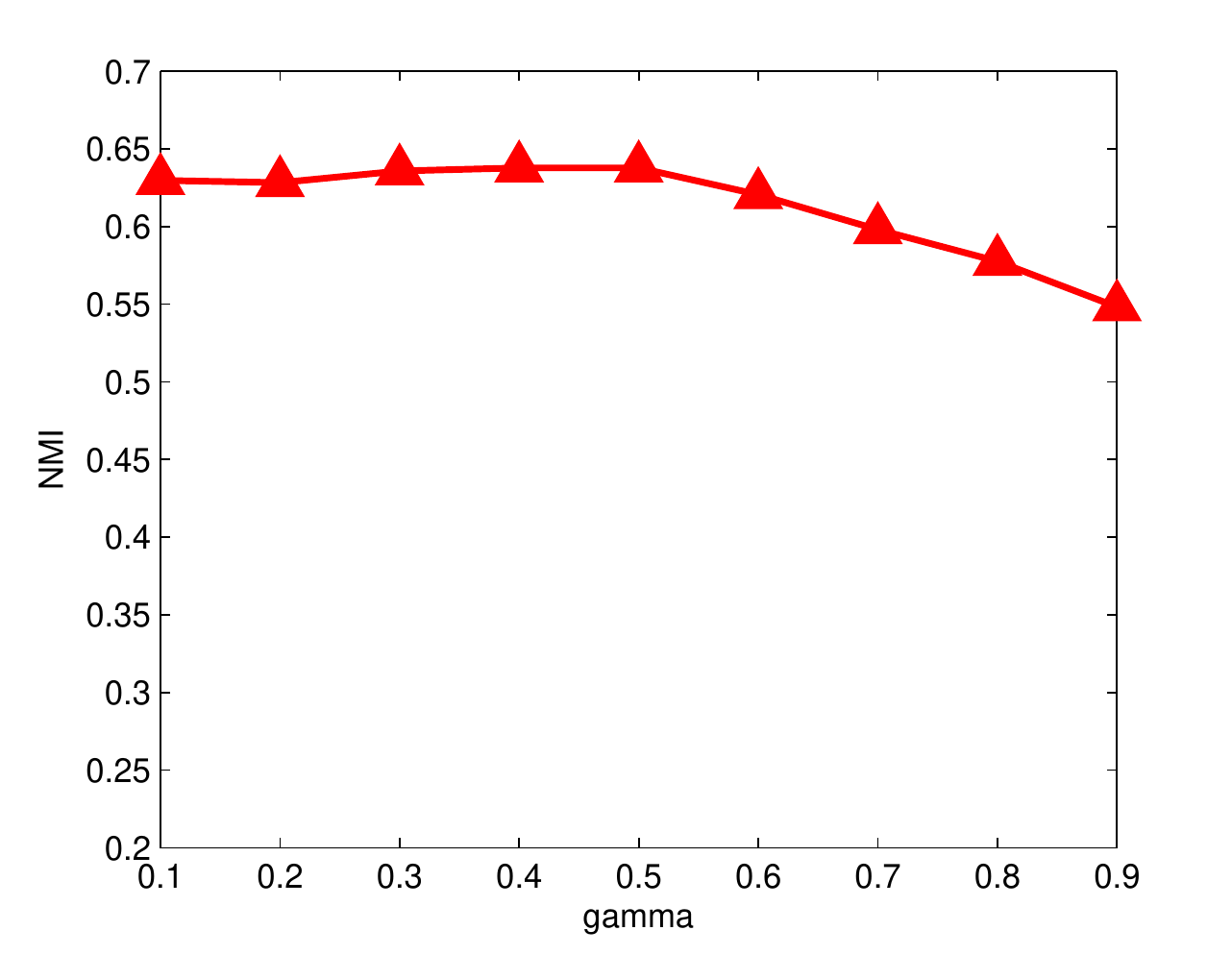}} &
\subfloat[Corel4]{\includegraphics[width=1.5in]{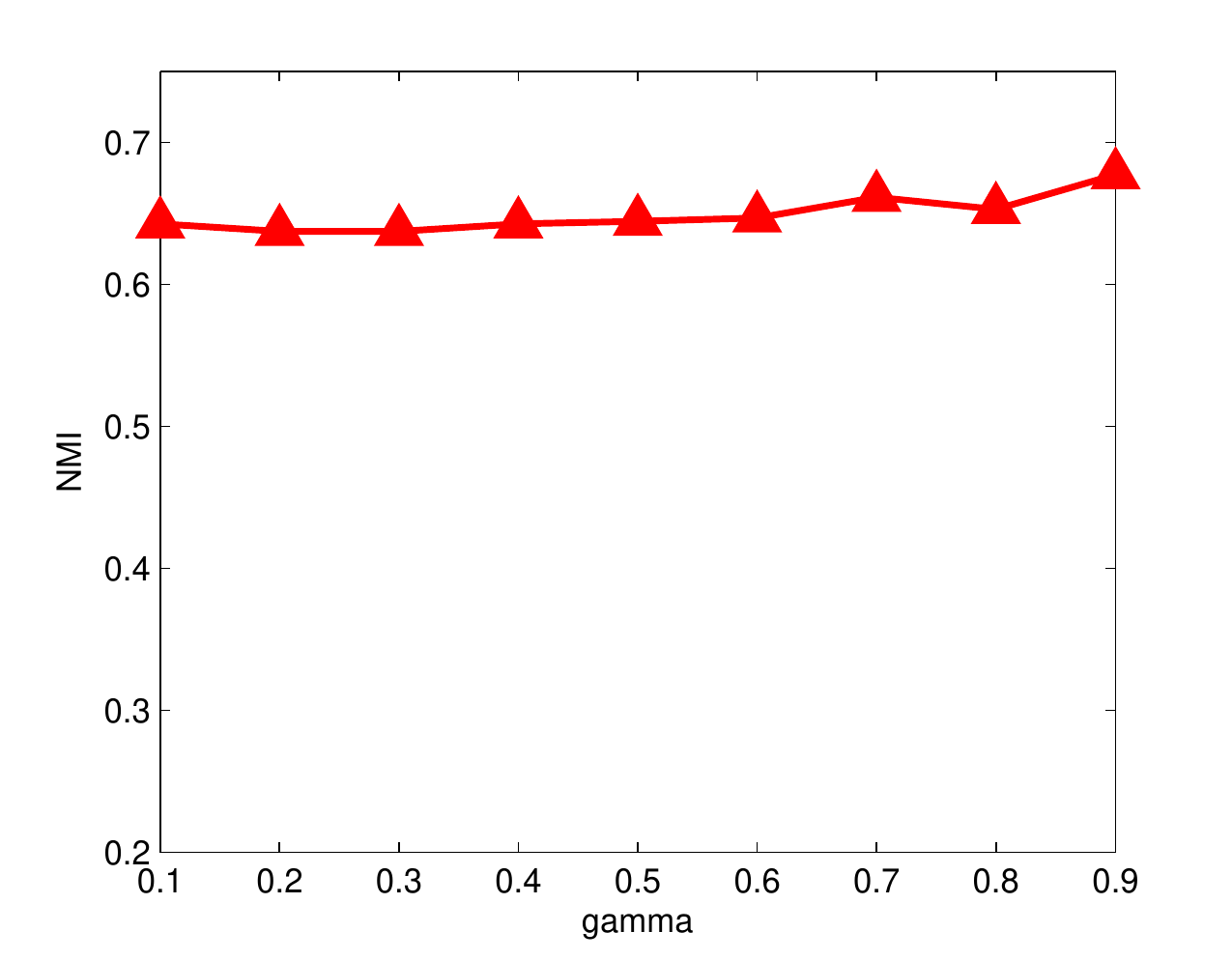}} \\
\subfloat[Corel5]{\includegraphics[width=1.5in]{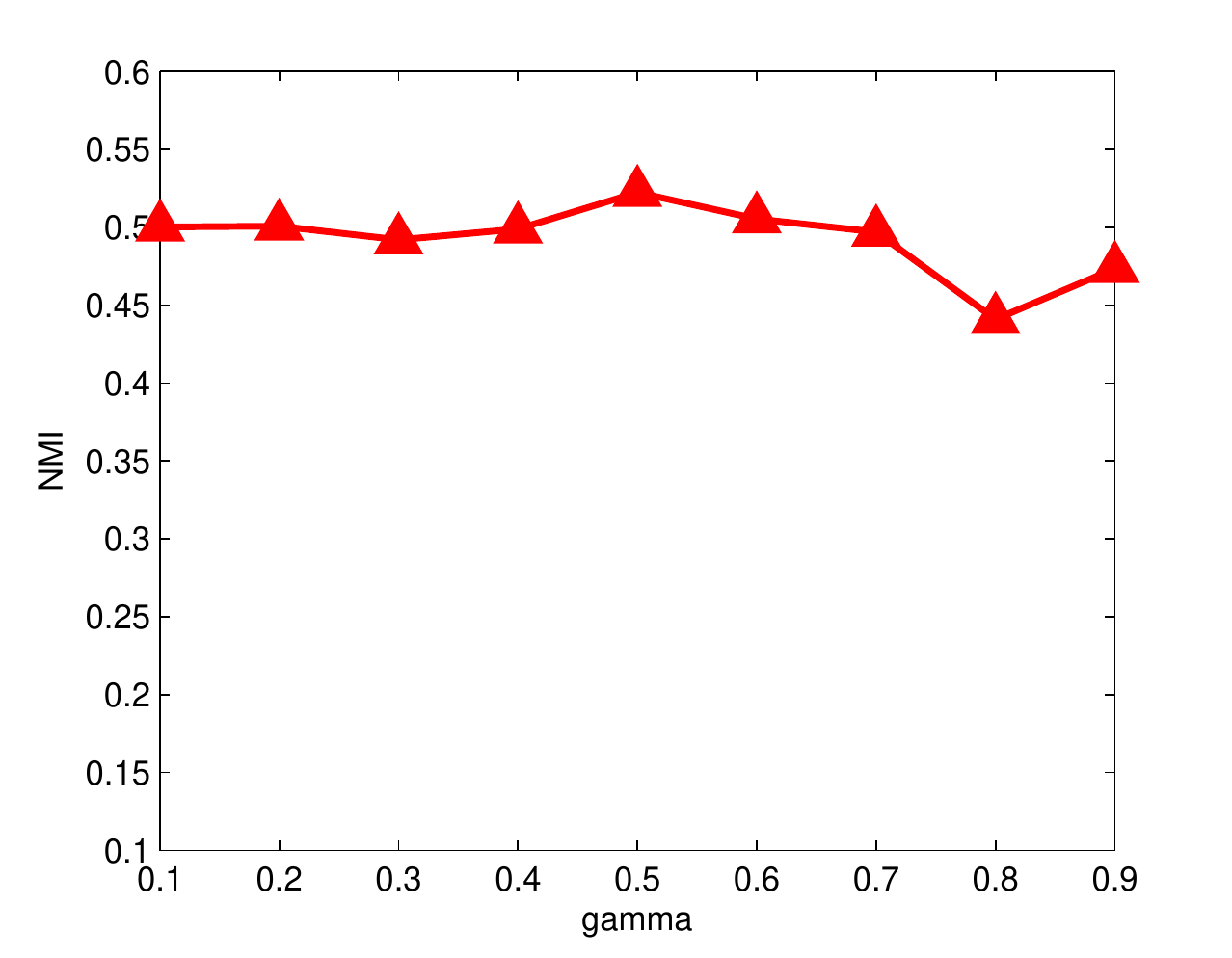}} &
\subfloat[3-Sources document]{\includegraphics[width=1.5in]{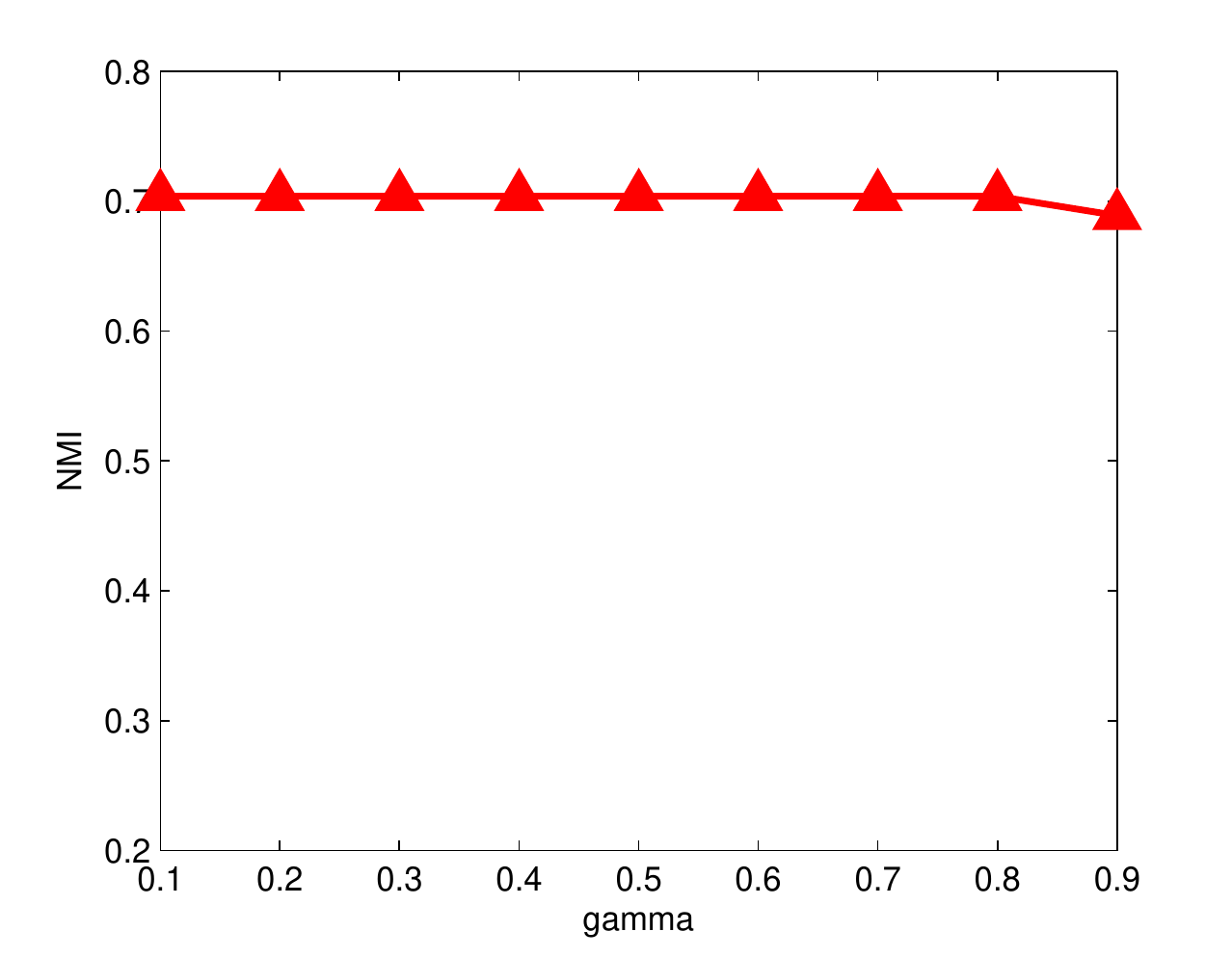}} &
\subfloat[Reuters multilingual data]{\includegraphics[width=1.5in]{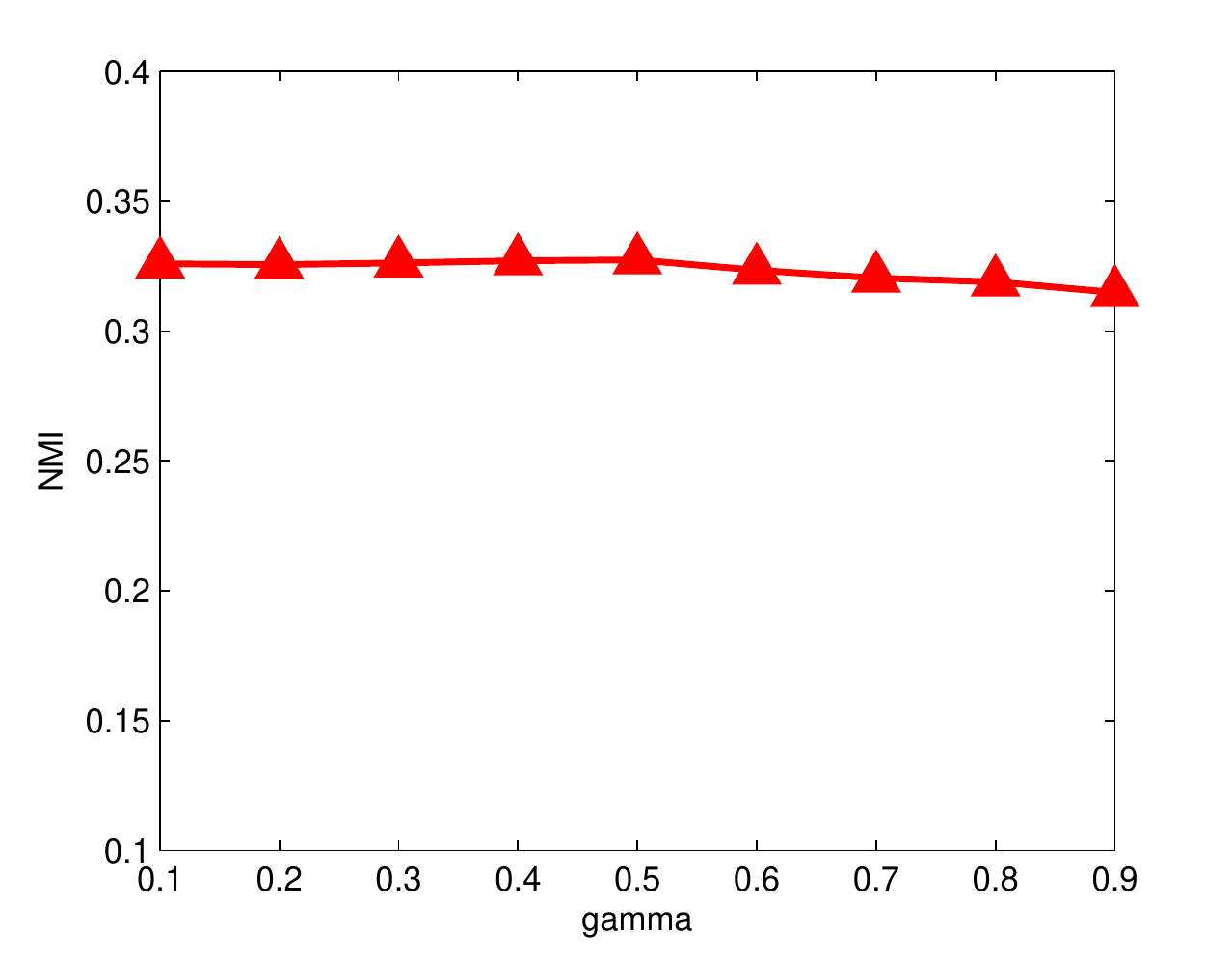}} \\
\end{tabular}
\caption{NMI results for each data set with different values of parameter $\gamma$}
\label{fig:paraGamma}
\end{figure}

\begin{figure}
\centering
\begin{tabular}{ccc}
\subfloat[Multiple features]{\includegraphics[width=1.5in]{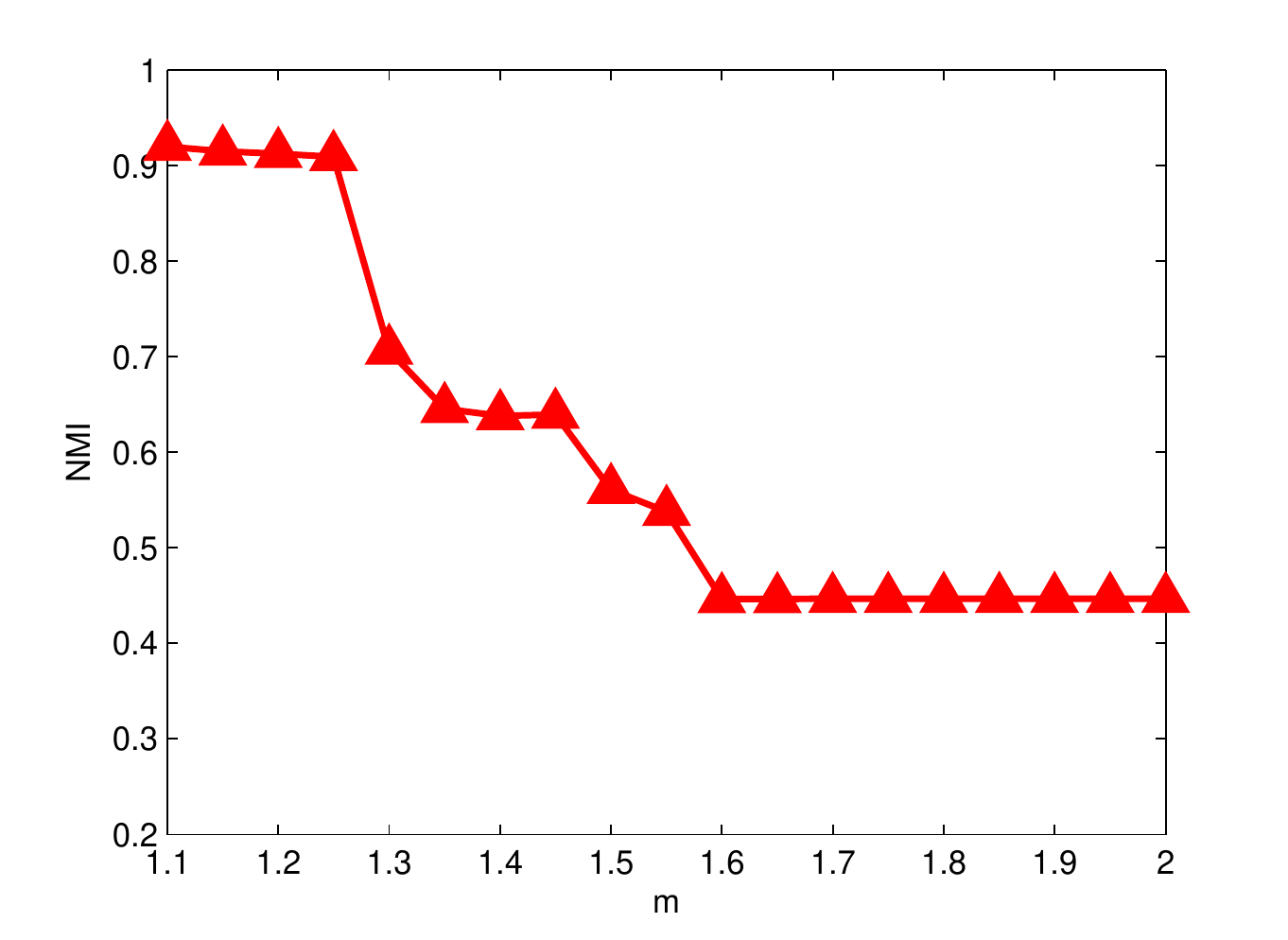}} &
\subfloat[Image segmentation]{\includegraphics[width=1.5in]{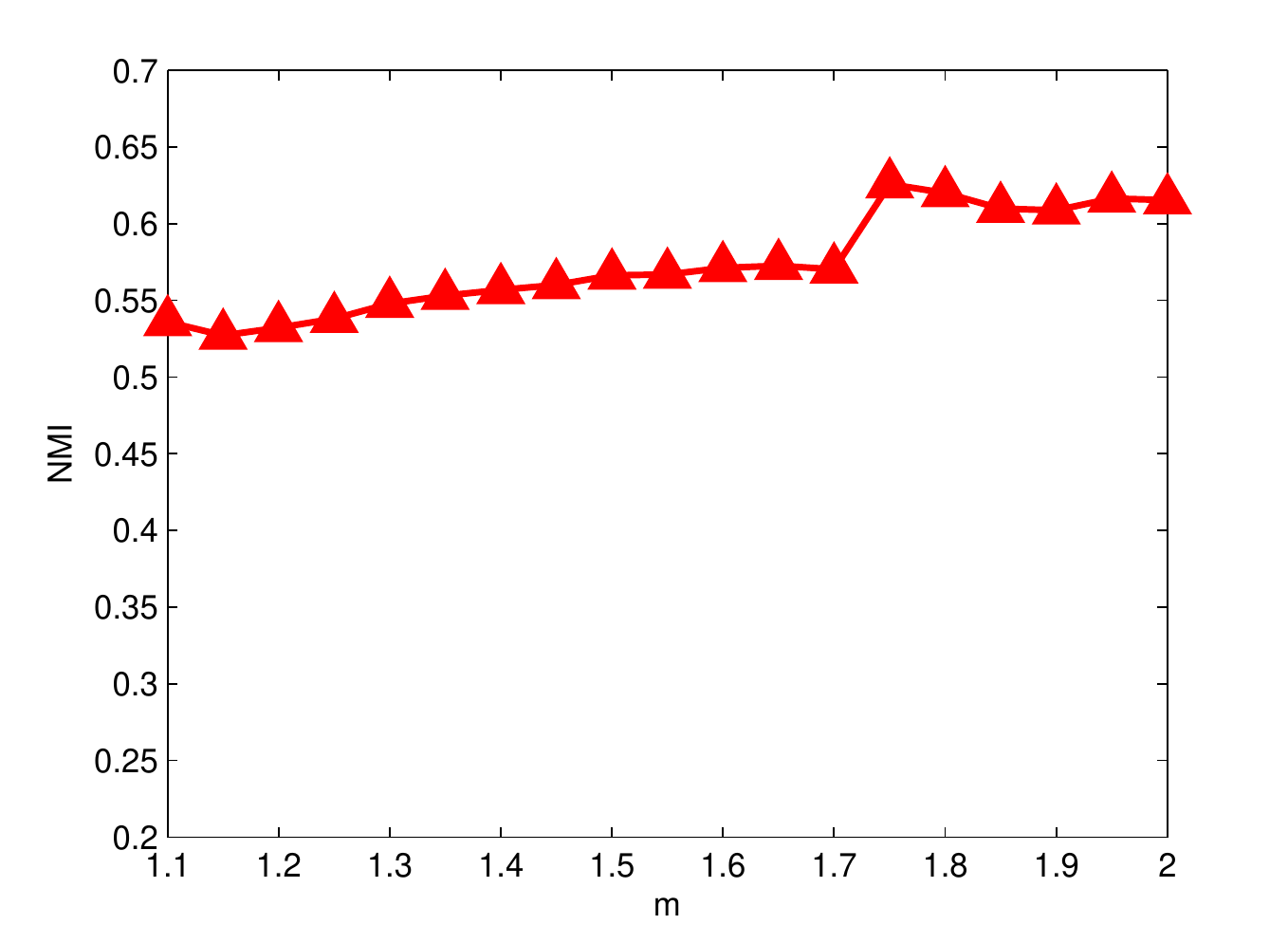}} &
\subfloat[Corel1]{\includegraphics[width=1.5in]{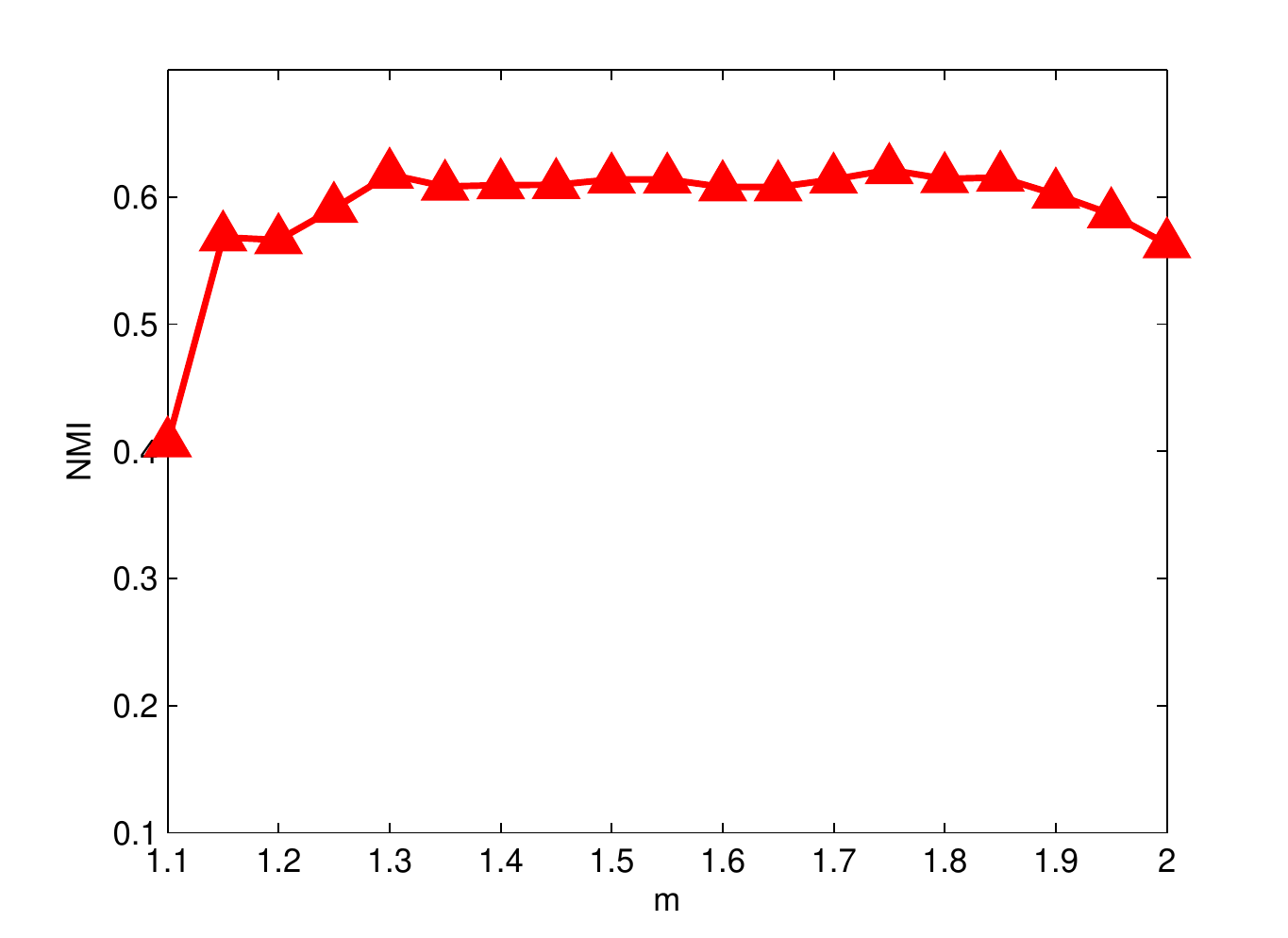}} \\
\subfloat[Corel2]{\includegraphics[width=1.5in]{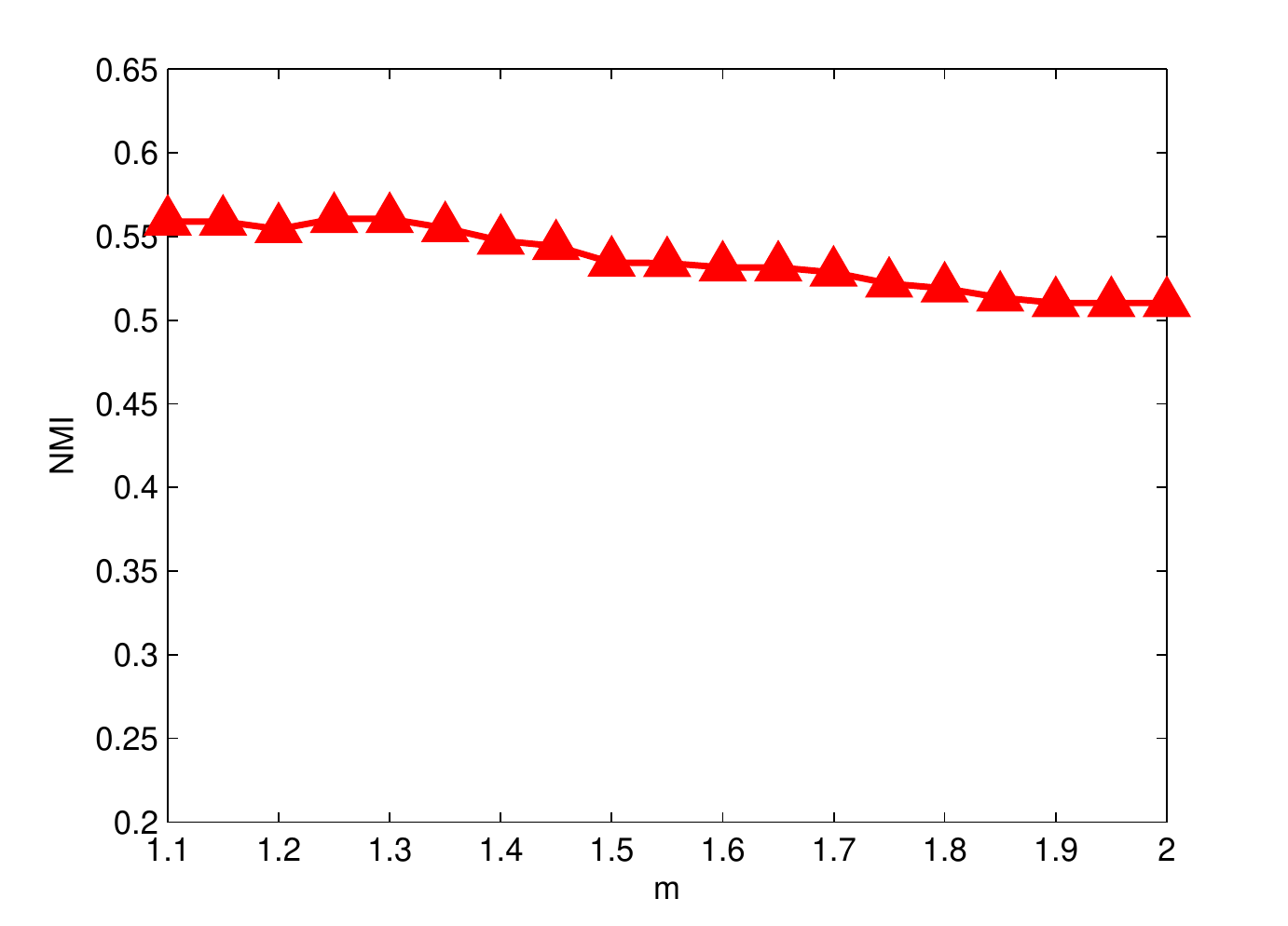}} &
\subfloat[Corel3]{\includegraphics[width=1.5in]{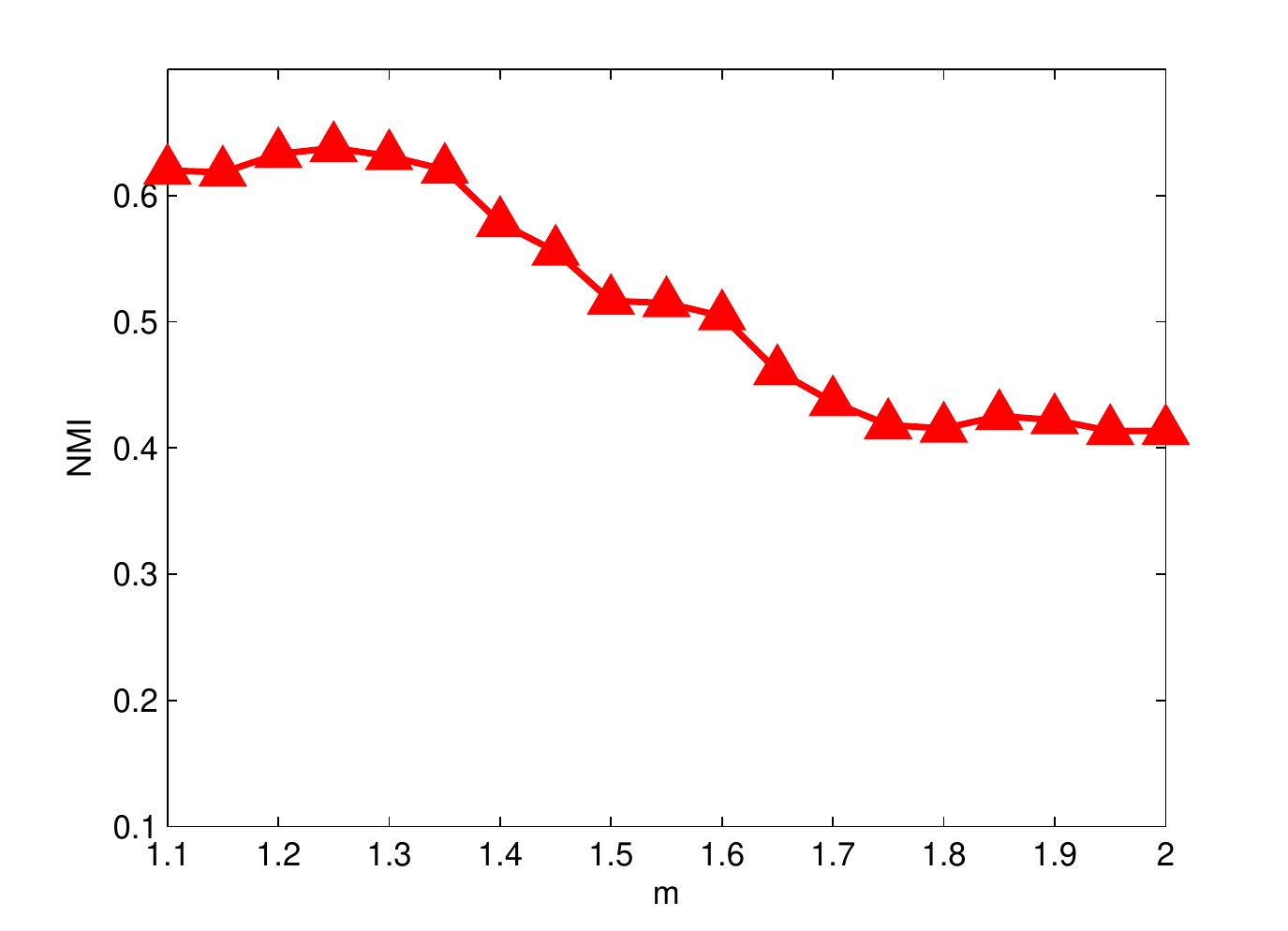}} &
\subfloat[Corel4]{\includegraphics[width=1.5in]{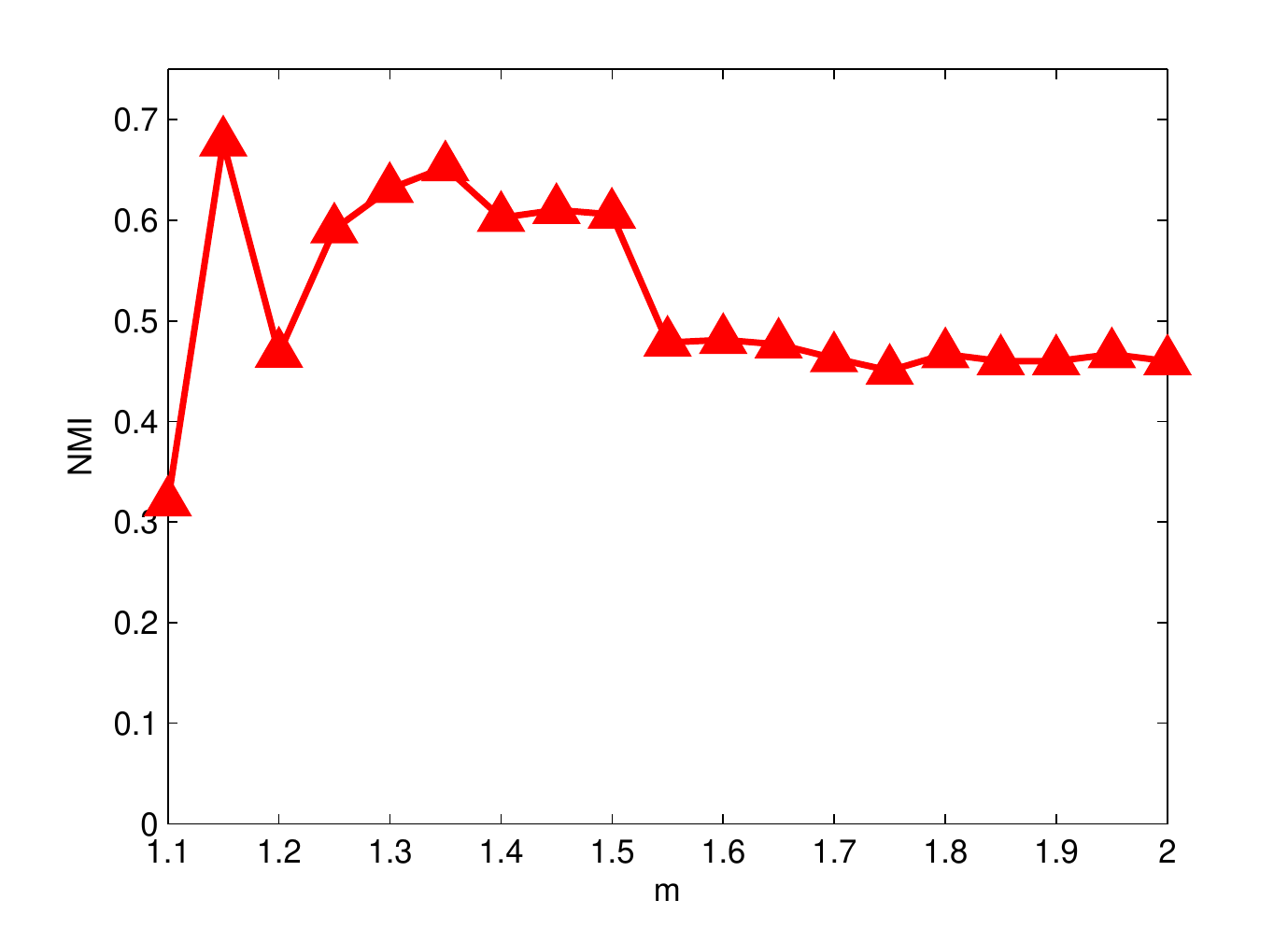}} \\
\subfloat[Corel5]{\includegraphics[width=1.5in]{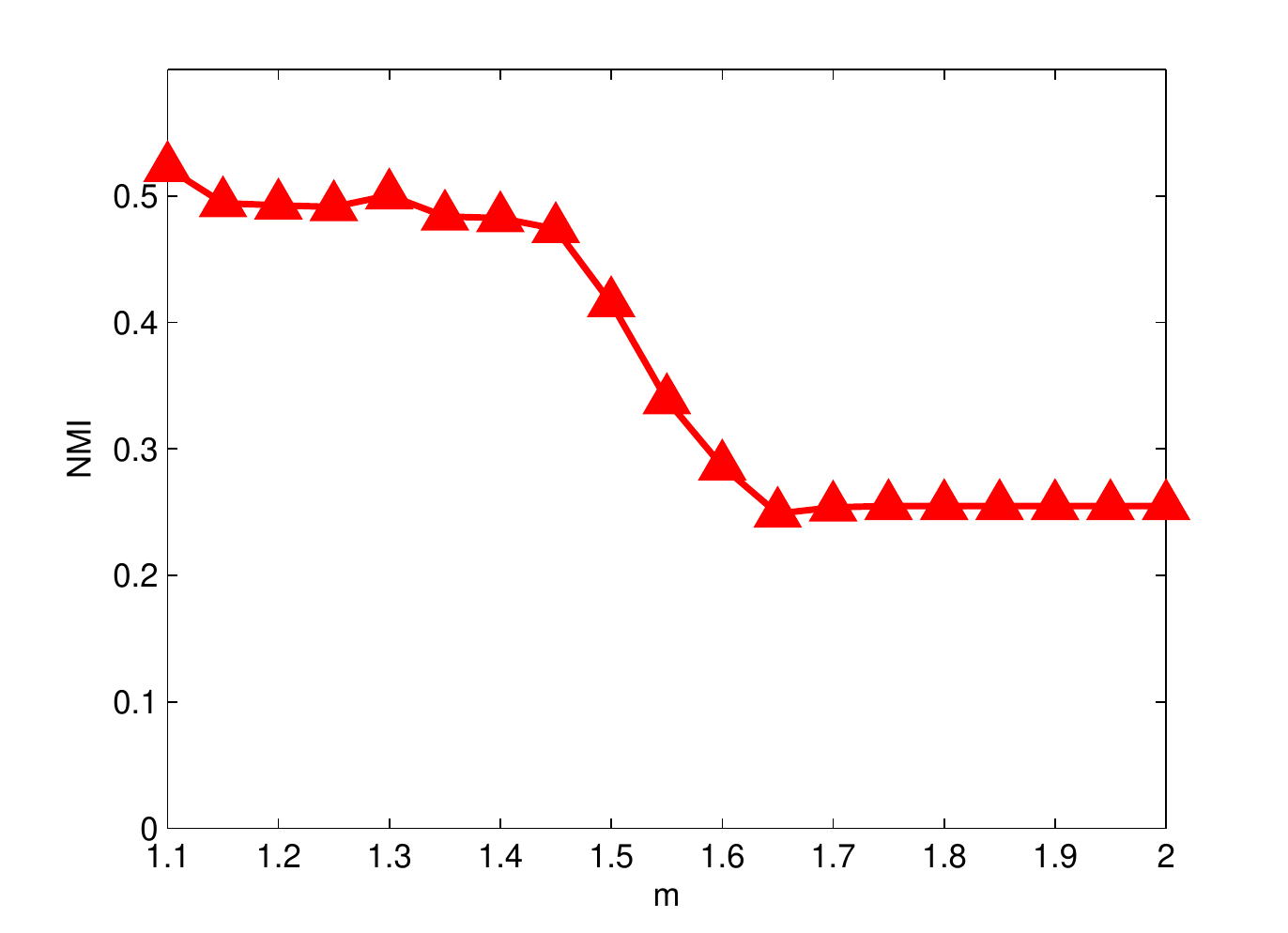}} &
\subfloat[3-Sources document]{\includegraphics[width=1.5in]{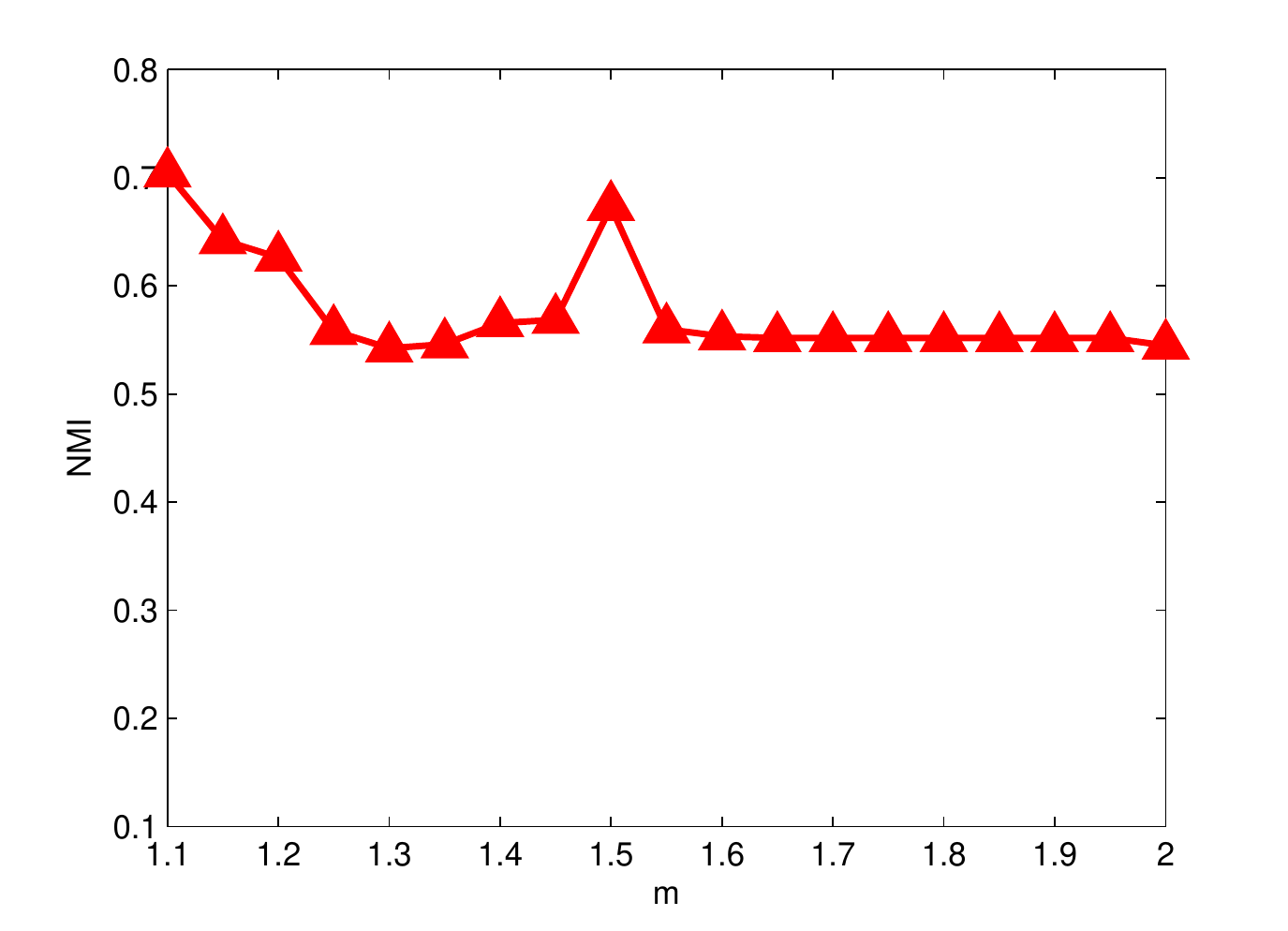}} &
\subfloat[Reuters multilingual data]{\includegraphics[width=1.5in]{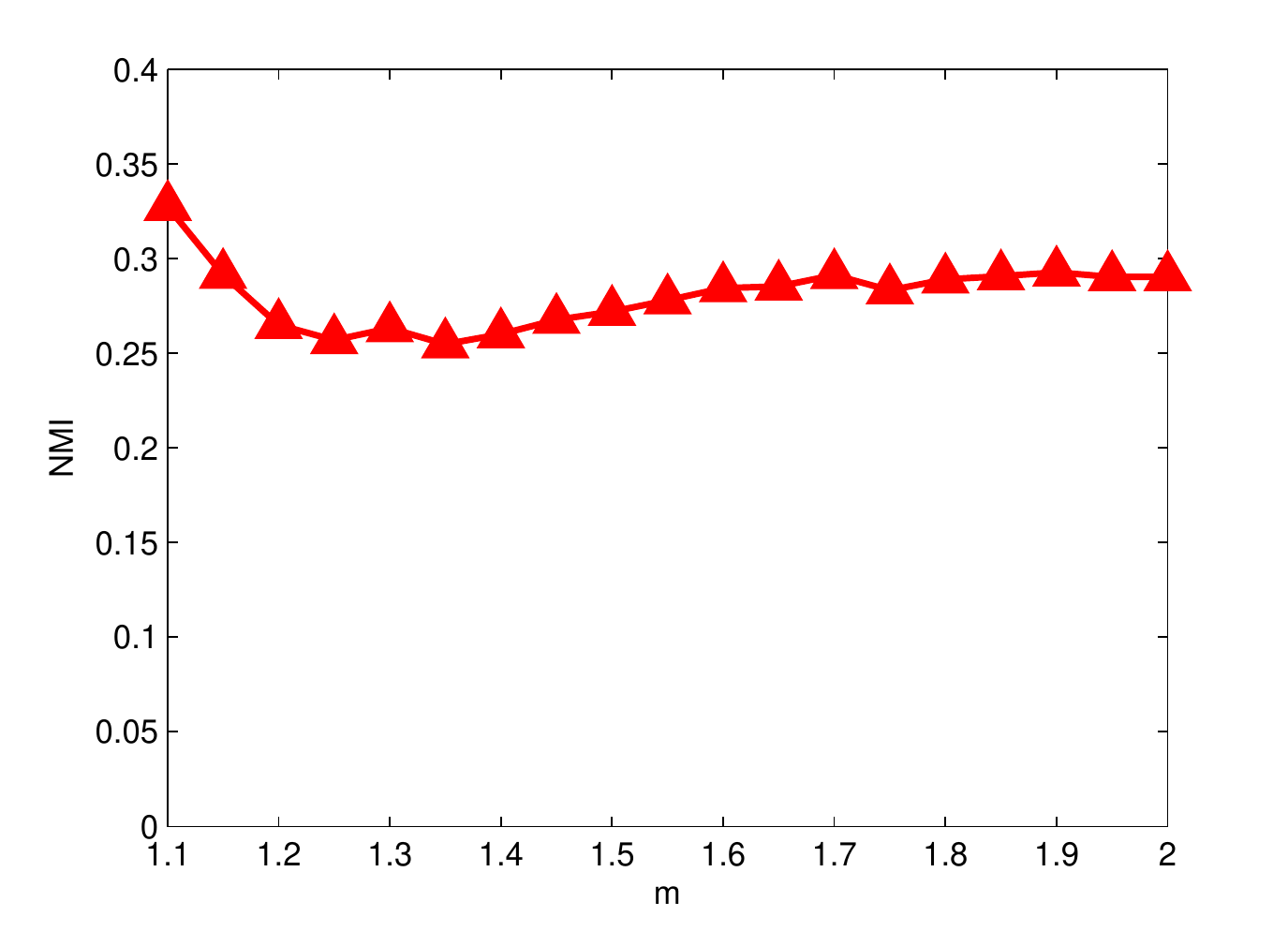}} \\
\end{tabular}
\caption{NMI results for each data set with different values of fuzzifier $m$}
\label{fig:paraM}
\end{figure}

The formulation of the objective function of MinimaxFCM has two parameters($\gamma$ and fuzzifier $m$) as shown in (\ref{equation:JMVFC}) and (\ref{equation:QP}). To show the impact of the two parameters on the performance of MinimaxFCM, we plot the NMI performance curve w.r.t. the parameters for each data set in Fig.~\ref{fig:paraGamma} and Fig.~\ref{fig:paraM}, respectively. Here we only show the NMI results, the results of accuracy and F-measure have a similar pattern. Fig.~\ref{fig:paraGamma} and Fig.~\ref{fig:paraM} are generated as follows. First, for Fig.~\ref{fig:paraGamma} the vaule of the fuzzifier $m$ which produces the results in Table.~\ref{table:mf}, Table.~\ref{table:is} and Table.~\ref{table:imagedataresults} is fixed.  Then, the NMI results is plotted w.r.t the parameter $\gamma$ with values from  [0.1 0.9]  with the step 0.1.  Fig.~\ref{fig:paraM} is plotted using the same method as Fig.~\ref{fig:paraGamma}, while the NMI results is plotted w.r.t the fuzzifier $m$ with values from [1.1 2] with the step 0.05. As shown in Fig.~\ref{fig:paraGamma} and Fig.~\ref{fig:paraM}, the NMI results are not very sensitive to the parameter $\gamma$  for each data set. Compared to $\gamma$, the NMI results are more sensitive to the fuzzifier $m$. We observe that the values of fuzzifier $m$ are always in the range  of [1.1 1.7] when the best NMI results are achieved. Moreover, for document data a smaller fuzzifier $m$ achieves better NMI results. Therefore, we recommend to set $\gamma$  to a value from [0.1 0.9] and $m$ from [1.1 1.7] in practice. In addition, if the data set is document data, a smaller $m$ is more suitable.

\section{Conclusion}
We have proposed a new multi-view fuzzy clustering approach called MinimaxFCM for multi-view data analysis, and apply MinimaxFCM on seven image data sets and two document data sets to demonstrate its effectiveness and potential. MinimaxFCM processes multi-view data based on the minimax optimization and the standard FCM to get the harmonic consensus clustering results. The maximum of the weighted cost of each view is minimized. Experimental results show that MinimaxFCM outperforms related multi-view clustering algorithms with more accurate clustering results. Moreover, the time complexity of MinimaxFCM is similar to FCM which indicates that MinimaxFCM has a great potential to be used for large multi-view data clustering. In the future, MinimaxFCM may be further extended to handle the scenario where the entire data set is too large to be stored in the memory for clustering tasks.
%
%

\section*{References}
\bibliography{reference}

\end{document}